\documentclass[10pt, conference, compsocconf]{IEEEtran}

\usepackage{enumerate}
\usepackage{color}
\usepackage{colortbl} 
\usepackage[pdftex]{graphicx}
\usepackage{graphicx}
\usepackage[bookmarks=false]{hyperref}
\usepackage{subcaption}
\usepackage{mathtools} 
\hypersetup{colorlinks=true,linkcolor=black,citecolor=black,filecolor=black,urlcolor=blue}

\newcommand{\todo}[1]{}

\newcommand\Longitude{{Longitude}}
\newcommand\Latitude{{Latitude}}
\newcommand\DoorNumberlabel{{Door\_number}}
\newcommand\Streetlabel{{Street\_label}}
\newcommand\AptUnitlabel{{Apartment\_number}}
\newcommand\Citylabel{{City}}
\newcommand\IdProvince{{Province}}
\newcommand\PostalCodelabel{{Zip\_code}}
\newcommand\IdZone{{Zone}}
\newcommand\IdTruck{{Truck}}
\newcommand\IdDriver{{Driver}}
\newcommand\RoadOrder{{Route\_order}}
\newcommand\PickupDateTime{{Planned\_Service\_Date\_Time}}
\newcommand\DistanceFromPrevious{{Distance\_from\_Previous\_Stop}}
\newcommand\TimeFromPrevious{{Time\_from\_Previous\_Stop}}
\newcommand\TimeWindowPickupStartTime{{Time\_Window\_Start\_Time}}
\newcommand\TimeWindowPickupEndTime{{Time\_Window\_End\_Time}}
\newcommand\TimeWindowSize{{Time\_Window\_Size}}
\newcommand\StartSlack{{Start\_Slack}}
\newcommand\EndSlack{{End\_Slack}}
\newcommand\IsAutomaticCallAllowed{{Automatic\_Calls}}
\newcommand\IdOutboundCallStatus{{Call\_Status}}
\newcommand\IdOutboundCallAttemptResult{{Detailed\_Call\_Status}}
\newcommand\WeekofYear{{Week\_of\_Year}}
\newcommand\Timeofday{{Time\_of\_Day}}
\newcommand\Day{{Day\_of\_Week}}
\newcommand\IdTaskType{{Service\_Type}}
\newcommand\IdCompany{{Retailer}}
\newcommand\Volumecf{{Item\_Volume}}
\newcommand\Weightkg{{Item\_Weight}}
\newcommand\IdManufacturer{{Item\_Manufacturer}}
\newcommand\EstimatedJobTime{{Estimated\_Service\_Time}}

\definecolor{Rowcolor}{gray}{0.9}
\definecolor{rowred}{rgb}{1,0.8,0.8}
\definecolor{rowgreen}{rgb}{0.8,1,0.8}

\begin{document}

\title{Data models for service failure prediction in supply-chain networks\vspace*{-0.3cm}}

\author{Monika Sharma$^1$, Tristan Glatard$^1$, \'Eric G\'elinas$^2$, Mariam Tagmouti$^2$, Brigitte Jaumard$^1$\\
  $^1$Department of Computer Science and Software Engineering, Concordia University \\ $^2$ ClearDestination Inc \\ Montr\'eal, Qu\'ebec, Canada \vspace*{-0.3cm}}

\maketitle

\begin{abstract}
We aim to predict and explain service failures in supply-chain 
networks, more precisely among last-mile pickup and delivery services 
to customers. We analyze a dataset of 500,000 services using
(1) supervised classification with Random Forests, and (2) Association 
Rules. Our classifier 
reaches an average sensitivity of 0.7 and an average specificity of 0.7 for the 
5 studied types of failure. Association Rules reassert
the importance of confirmation calls to prevent failures due to 
customers not at home, show the importance of the time window size, 
slack time, and geographical location of the customer for the other 
failure types, and highlight the effect of the retailer company on 
several failure types. To reduce the occurrence of service failures, 
our data models 
could be coupled to optimizers, or used to define counter-measures to 
be taken by human dispatchers.
\end{abstract}

\section{Introduction}

Service failures are pervasive in 
supply-chain networks, with important consequences on their 
cost-efficiency and customer experience. We aim at predicting and 
explaining the cause of such failures, focusing on the last-mile pickup 
and delivery of items at customer locations. Such services are 
planned by optimizers solving some variations of the Vehicle-Routing 
Problem, in our case the Pickup and Delivery Problem with Time Windows 
(PDPTW~\cite{ropke2009branch}). Solutions consist 
of \emph{routes}, eventually associated to a specific truck and driver, 
defined as a sequence of \emph{stops} associated with customer 
locations, where \emph{services} are delivered, with time-window, capacity and 
precedence constraints. Routes start and end at dispatch 
centers, where trucks are loaded.

We aim at predicting service failures that occur after the route was 
planned, such as customer not at home, services rescheduled by dispatch 
center, and service refused or canceled by customer. Such predictions 
could be leveraged by optimizers to find routes with reduced 
amounts of failures, for instance by adjusting the slack time to 
maximize the probability that customers will be at home at the time of 
the service. They could also be used directly by human planners, for 
instance to increase the amount of confirmation calls to customers in 
case of high failure probability. 

Current approaches to solve the pick-up and delivery problem with time 
windows include heuristics~\cite{rop06}, 
meta-heuristics~\cite{nanry2000solving} and exact 
methods~\cite{ropke2009branch}. To the best of our knowledge, they do not
take service failures into account. Instead, the solutions 
suggested in the literature to reduce service failure in last-mile 
problems include Collection Delivery Points~\cite{song2009addressing}, 
reception boxes, and delivery boxes~\cite{punakivi2001solving}. While 
these methods greatly improve service efficiency, they also reduce 
customer satisfaction by not delivering items directly to their homes. 
Instead, we aim at (1) predicting failure probabilities, (2) identifying 
factors that predict failures, and (3) suggesting counter-measures to avoid 
failures.

We break down the problem of failure prediction into multiple, 
independent supervised classification problems. Given a failure type, 
our goal is to predict if a stop will fail with this type. We are 
also looking for association between stop features and failure, to 
provide specific ranges
of values that lead to increased or reduced failure rates.

We build our classifier using Random Forests~\cite{breiman2001random}, 
as they are one of the most successful and frequently used method for 
supervised classification. Random Forests belong to ensemble learning 
methods: they combine the 
predictions of multiple decision trees built 
from randomly-selected features, to reduce overfitting. Random 
Forest 
also provide measures of feature importance, which helps interpret 
classification results. To provide further insights on the failure causes,
we extract Association Rules between categorized feature values and 
specific failure types. 

We analyze a large dataset of more than 500,000 pick-up and delivery 
services that occurred in Canada over a period of 6 months, among which a few 
percents failed. The dataset is strongly imbalanced toward successful 
services, as it is commonly the case in failure analysis. This is an 
issue for classifiers as they tend to focus on the majority class, 
leading to poor sensitivity. To address this issue, we leverage resampling methods 
to either undersample the majority 
class~\cite{mani2003knn} or oversample the minority class~\cite{chawla2002smote}.

This paper makes the following contributions:
\begin{itemize}
\item We apply Random Forests to the prediction of stop failures in the 
pickup and delivery problem with time windows and precedence 
constraints.
\item We apply Association Rules to the identification of factors 
predicting failures types.
\item We analyze a dataset of 500,000 pickup and delivery services, on 
which we quantify the performance of the classifier and highlight 
feature ranges leading to increased failure probabilities, for 5 
failure types.
\end{itemize}

Section ~\ref{sec:method} presents the dataset and methods used, and 
Section~\ref{sec:results} details the results. The paper closes on a 
discussion on the performance of the Random Forest classifier, on the 
relevance of resampling methods, and on possible counter-measures to 
reduce failure rates.

\section{Dataset and methods}
\label{sec:method}
\subsection{Dataset}

\subsubsection{Presentation}

We extracted a dataset from the database of ClearDestination Inc, a 
Montreal-based company providing services for supply-chain optimization.
The dataset contains 523,643 pickup and delivery services grouped in 
183,872 stops, scheduled between September 2017 and February 2018 in 
Canada. Table~\ref{table:used-features} shows the 
features describing the stops and their services. Features describe the 
customer location, position of the stop in the route produced by the 
optimizer, phone call status (phone calls are placed to the customer at 
specific times before the service), date, and service characteristics. 
Some features may be correlated: for instance, features describing geographical location 
are strongly interdependent.

\begin{table*}[h]
\centering
\begin{subfigure}{\textwidth}
\centering
\begin{tabular}{@{}|llll|@{}}
\hline
Id & \multicolumn{3}{c|}{Feature name \hfill \hfill Description \hfill \hfill \hfill \hfill Type}\\
\hline
\rowcolor{Rowcolor}
\multicolumn{4}{|c|}{Customer geographical location (C)}\\
C1 & \Longitude                   & Longitude of the customer location                                                    & Numerical\\
C2 & \Latitude                    & Latitude of the customer location                                                     & Numerical\\
C3 & \DoorNumberlabel             & Door number in the customer address                                                  & Categorical\\
C4 & \Streetlabel                 & Street number in the customer address                                                 & Categorical\\
C5 & \AptUnitlabel                & Apartment number in the customer address                                             & Categorical\\
C6 & \Citylabel                   & City in the customer address                                                         & Categorical\\
C7 & \IdProvince                  & Province of the customer location                                                  & Categorical\\
C8 & \PostalCodelabel             & Postal code in the customer address                                                  & Categorical\\
C9 & \IdZone                      & Zone identifier of the customer location                     & Categorical\\
\hline
\rowcolor{Rowcolor}
\multicolumn{4}{|c|}{Route schedule (R)}\\
R1 & \IdTruck                     & Vehicle allocated to the service                             & Categorical\\
R2 & \IdDriver                    & Driver identifier                                            & Categorical\\
R3 & \RoadOrder                   & Order of the stop in the route                    & Numerical\\
R4 & \PickupDateTime              & Service time planned by the optimizer               & Numerical\\
R5 & \DistanceFromPrevious        & Distance from the previous stop in the route (km)                         & Numerical\\
R6 & \TimeFromPrevious            & Time from the previous stop in the route (min)                             & Numerical\\
R7 & \TimeWindowPickupStartTime   & Start time of the service time window (min since 00:00)      & Numerical\\
R8 & \TimeWindowPickupEndTime     & End time of the service time window (min since 00:00)                          & Numerical\\
R9 & \TimeWindowSize              & Service time window span   (min)                                  & Numerical\\
R10 & \StartSlack                 & \parbox{10cm}{Time difference between pickup time and start of time window (min). May be negative when the service is performed before the time window starts.} & Numerical\\
R11 & \EndSlack                   & \parbox{10cm}{Time difference between end of time window and pickup time (min). May be negative when the service is performed after the time window ends.}& Numerical\\
\hline
\rowcolor{Rowcolor}
\multicolumn{4}{|c|}{Phone calls (P)}\\
P1 & \IsAutomaticCallAllowed     & If automated calls are allowed (yes or no)                               & Categorical\\
P2 & \IdOutboundCallStatus       & Status of phone call to customer (completed: 5, failed: 6)           & Categorical\\
P3 & \IdOutboundCallAttemptResult & Detailed status of the call (answered: 2, voice mail: 3, other failures: 4-21)      & Categorical\\
\hline
\rowcolor{Rowcolor}
\multicolumn{4}{|c|}{Date and time (D)}\\
D1 & \WeekofYear                 & Week number of the year (1-52)                             & Categorical\\
D2 & \Timeofday                  & \parbox{10cm}{Time in the day (early morning: 1, morning:2, afternoon: 4, early evening: 5, evening: 6, late evening: 7, night: 8}& Categorical\\
D3 & \Day                        & Day of the week (1-7)                                        & Categorical\\
\hline
\end{tabular}
\caption{Stops}
\end{subfigure}
\begin{subfigure}{\textwidth}
\centering
\begin{tabular}{@{}|llll|@{}}
\hline
Id & \multicolumn{3}{|c|}{Feature \hfill \hfill Description \hfill \hfill \hfill \hfill Type}\\
\hline
S1 & \IdTaskType                & Service type (Pickup or Delivery)                            & Categorical\\
S2 & \IdCompany                 & Retailer identifier                                   & Categorical\\
S3 & \Volumecf                  & Volume of the item to be picked up or delivered (cf)                                & Numerical\\
S4 & \Weightkg                  & Weight of the item (lbs)                           & Numerical\\
S5 & \IdManufacturer            & Item manufacturer                                              & Categorical\\
S6 & \EstimatedJobTime          & Estimated service time (s)                                       & Numerical\\
\hline
\end{tabular}
\caption{Services}
\end{subfigure}
\caption{Features of stops and services (S)}
\label{table:used-features}
\end{table*}
\subsubsection{Data representation}

The dataset contains records where each stop is associated to one or 
more services, represented as a vector of variable size (3 services per 
stop on average). This is a problem since classifiers cannot work on 
spaces containing vectors of variable size. A solution could be to 
create one record for each service, and to replicate the 
stop features in all the services associated with the stop. However, 
such a replication would likely lead the classifier to overfit 
particular stops, and to rely, for instance, on the exact latitude and 
longitude of the stop to predict failures. This is not 
desirable in our case, since
we aim at predicting failures in situations more general than 
particular stop locations that may never re-occur.

\begin{figure}
\centering
\includegraphics[width=\columnwidth]{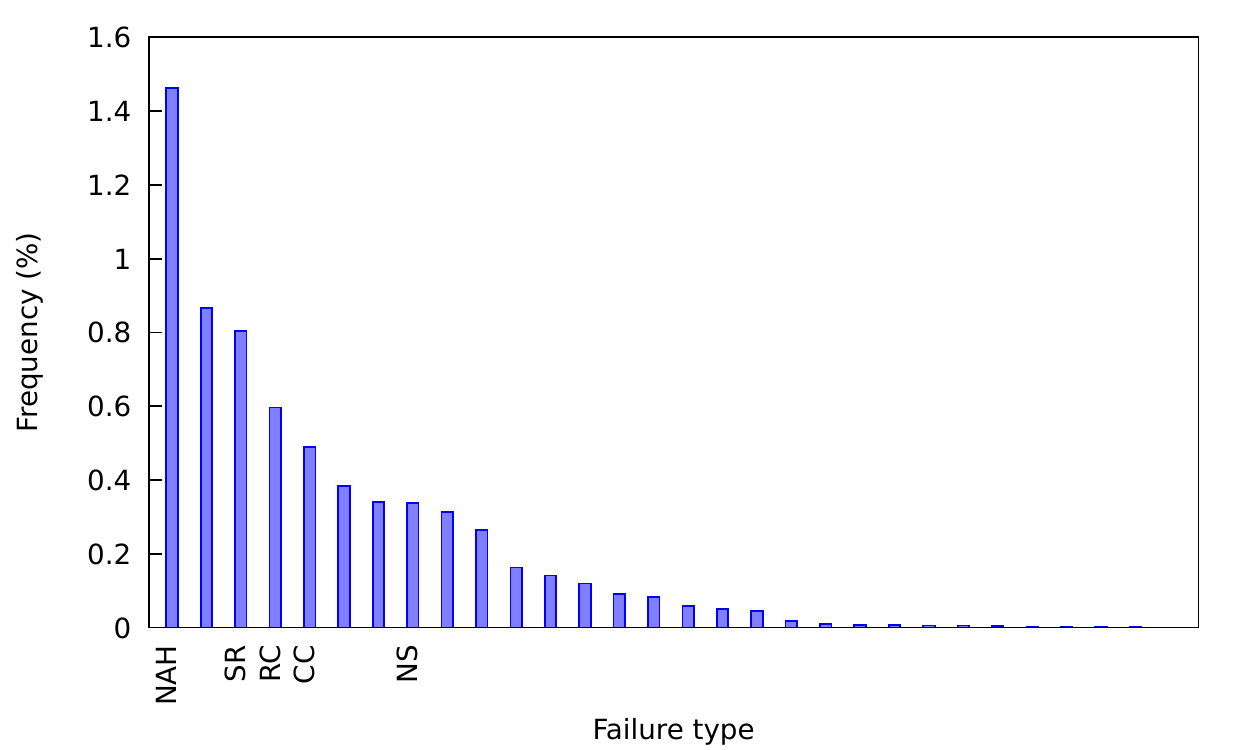}
\footnotesize
\caption{Distribution of failure types.}
\label{fig:failure-distribution}
\end{figure}

Instead, we aggregated the services of a particular stop in a ``master 
service", for which the value of categorical service features (\IdTaskType, 
\IdCompany{ }and \IdManufacturer) was determined as the most frequent 
value among services in the stop, and the value of numerical features 
(\Volumecf, \Weightkg{ }and \EstimatedJobTime) was the sum of the values 
among services in the stop. The status and failure type of the stop 
were set as the most frequent status and failure type among its 
services. The resulting aggregated dataset 
contains 183,872 stops with 6.68\% of failures. 
Figure~\ref{fig:failure-distribution} shows the failure distribution by 
failure type in the aggregated dataset.

\subsubsection{Failure types}

We focus on some of the most frequent failures, namely:
\begin{itemize}
\item Customer not at home (NAH, frequency: 1.46\%)
\item Stop rescheduled by dispatch center (SR, 0.80\%)
\item Refused by customer (RC, 0.60\%)
\item Canceled by customer (CC, 0.49\%)
\item Not in Stock (NS, 0.34\%)
\end{itemize}
Failure type Customer not at home (NAH) happens when the customer is not present at the 
time of the service. Failure type Stop rescheduled by
dispatch center (SR) may happen due to any unexpected event in the 
supply chain, for instance construction in the delivery area, or 
inbound delays at the dispatch center. Failure type Refused by 
customer (RC) means that the item was delivered to the customer's 
place, but the customer refused it, perhaps because it did not match 
their expectations. Failure type Canceled by customer (CC) means 
that the service was canceled at the customer's location, for instance 
because the customer did not have cash. Finally, failure type Not in 
stock (NS) means that the item was not in stock at the dispatch center 
on the day of delivery, which may happen when the information is unknown at the time of the planning.

The failure types present in the dataset may 
 not always be very accurately used. This is due to the 
 fact that they are set by different actors in the supply chain, among 
 which the planning company, the dispatcher and the drivers, who may 
 interpret the failure types differently. In particular, some failure types 
 may overlap, which may disturb the classification.

\subsubsection{Pre-processing}

We pre-processed the dataset to remove duplicate entries, to 
replace missing values with a constant (of -100), and to encode 
categorical features as numerical values. Only 3 features had missing 
values: \IdManufacturer{ }(S5, missing rate of 90.8 \%), \AptUnitlabel{ 
}(C5, 82.7\%), and \DoorNumberlabel{ }(C3, 8.6\%). In the remainder, 
results involving \IdManufacturer{ }and to some extend 
\DoorNumberlabel{ }should be interpreted carefully due to 
the high missing rate. Missing apartment numbers, however, make sense 
as many addresses simply do not include an apartment number.

\subsubsection{Dataset imbalance}

We used the following 3 methods to deal with dataset imbalance. 
First, we oversampled the minority class using Synthetic Minority 
Oversampling TEchnique (SMOTE~\cite{chawla2002smote}). SMOTE generates 
synthetic entries in the minority class, here the class of failed 
stops, using random linear combinations of existing failed stops. We 
applied SMOTE-regular using 2 nearest neighbors, and an oversampling 
ratio equal to the ratio between the number of elements in the majority 
class and the number of elements in the minority class (so that the 
resulting dataset is evenly distributed among both classes).

We also evaluated NearMiss, a method that undersamples the majority class
using k-nearest neighbours~\cite{mani2003knn}. We used NearMiss-3, which, for every
element in the minority class, determines its nearest neighbors and keeps only
the furthest ones. We applied this method with 3 nearest neighbors.

Finally, we assessed a straightforward random undersampling of the 
majority class, as we suspected that SMOTE and NearMiss would be 
disturbed by their use of the Euclidean distance to determine nearest 
neighbors, which is questionable in our dataset.

\subsection{Classification using Random Forest}
\label{sec:random-forest}
We trained Random Forest classifiers on the aggregated dataset, using a 
training ratio of 80\% and 5-fold cross validation. We used Random 
Forests as implemented in scikit-learn version 0.19.1. 

To find suitable values of the Random Forest parameters, we performed a 
grid search on a training dataset with different numbers of 
estimators, maximum depths of a tree, split criteria and minimum 
numbers of samples to split on a node. Table~\ref{table:rf-params} shows the selected
parameter values.
\begin{table*}
\centering
\begin{tabular}{@{}|llcc|@{}}
\hline
Parameter    & Description & Selected value & Grid search \\
\hline
\# Estimators       & Number of trees in the random forest & 100   & 10, 50, 100, 200, 500\\
Max depth           & Maximum depth of the trees            & 6    & 5, 6, 7, 8, 9, 10, 20, 50 \\
Split criterion     & Impurity criterion used in tree nodes & Gini &Gini, Entropy             \\
Max features       & Number of features to consider in each split & $\sqrt{n\_features}$ & -- \\
Min samples to split & Minimum number of samples required to split a node     & 10 & --   \\
Min sample in leaf   & Minimum number of samples required in a leaf node         & 5  & --    \\
OOB Score             & Out of bag error                                 & True  & --\\
\hline
\end{tabular}
\caption{Parameter settings used in Random Forest}
\label{table:rf-params}
\end{table*}

We also captured feature importance in the Random Forest 
classification, as reported by scikit-learn. It is defined by the 
scikit-learn developers\footnote{\url{https://stackoverflow.com/questions/15810339/how-are-feature-importances-in-randomforestclassifier-determined}}, 
as the \emph{total decrease in node impurity (weighted by the 
probability of reaching that node, which is approximated by the 
proportion of samples reaching that node) averaged over all trees of 
the ensemble}. However, feature importance provides limited insights to 
interpret classification results. One of the reasons is that 
correlation among a group features reduces the mean importance in this 
group of features, as discussed in ~\cite{genuer2010variable}. To 
further interpret the results, we also extracted Association Rules as 
explained hereafter.

\subsection{Association Rules}
\label{sec:association-Rules}
We extracted Association Rules from the aggregated dataset, to 
check the consistency of classification results and to provide further 
insights on their interpretation. To do so, we categorized numerical 
features into deciles, and we represented stops with vectors 
containing (1) such categorized features, (2) the initial categorical 
features, and (3) a binary feature representing the stop status 
(success or failure type).  An Association Rule, written as 
``antecedent $\Rightarrow$ consequent", consists of two tuples, an 
antecedent and a consequent ~\cite{agrawal1994fast}. We focus on the rules where the 
consequent is a singleton containing a stop status. To represent 
features in the antecedent, we postfix numerical features with 
\texttt{\_Dx}, to indicate that the value is in the 
\texttt{x}$^{\mathrm{th}}$ decile, and categorical features with 
\texttt{\_Vx}, to indicate that the value is \texttt{x}. A hypothetical 
example of rule is:
\[
   \mathrm{Start Slack\_D3, Day\_V4} \Rightarrow \mathrm{FAIL\_NAH},
\]
which measures the association between failure type ``Customer not at 
home", and stops where \StartSlack{ }is in the third decile and Day has 
value 4. It should be noted that Association Rules provide a measure of 
co-occurrence rather than causality. 

 To measure the relevance of a rule $x \Rightarrow y$, we define its 
 \emph{interest ratio} (IR) by comparing the frequencies of tuple x in 
 the complete dataset $C$ versus in the set $F$ of failed stops (the 
 stops that contain $y$):
\[
\mathrm{\phi}=\frac{\mathrm{sup}_F(x)}{|F|}\frac{|C|}{\mathrm{sup}_C(x)},
\]
where $\mathrm{sup_S(x)}$, the support of tuple $x$ in set $S$, is the 
number of occurrences of $x$ in $S$. We focus on the cases where 
$\mathrm{sup}_F(x) \neq 0$, which gives $\phi \neq 0$.
Then we define the interest ratio as follows:
\[
\mathrm{IR}(x \Rightarrow y)=\max\left( \phi, \frac{1}{\phi}\right).
\]
The interest ratio measures the effect of $x$ on the probability to fail 
with type $y$. Another way to understand it is to express
its relation to the failure probability conditional to the presence of $x$:
\[
P(y|x) = \frac{\mathrm{sup}_F(x)}{\mathrm{sup}_C(x)},
\]
which gives:
\[
P(y|x) = \mathrm{IR}(x \Rightarrow y) P(y) \quad (\phi \geq 1),
\]
or:
\[
 P(y|x) = \frac{P(y)}{\mathrm{IR}(x \Rightarrow y)} \quad (\phi \leq 1).
\]

We also compute the \emph{confidence} of rule $x \Rightarrow y$ with 
the usual definition:
\[
\mathrm{conf}(x \Rightarrow y)=\frac{\mathrm{sup}_C( x \cup y)}{\mathrm{sup}_C(x)}.
\]
The following relation should finally be noted:
\[
\mathrm{conf}(x \Rightarrow y)=\phi\frac{|F|}{|C|}.
\]

We are looking for rules with high interest ratio (large or small 
$\phi$), and high frequency in the failed set ($\mathrm{sup}_F(x) \geq s$, where $s$ is the desired support threshold in $F$). We find them
using the following approach:
\begin{enumerate}
\item Find the set $I$ of items $x$ in $F$ s.t. $\mathrm{sup}_F(x)\geq s$.
\item For every $x$ in $I$, compute $\mathrm{sup}_C(x)$. 
\end{enumerate}
We perform step 1 using the FP-growth algorithm~\cite{han2000mining}, 
as implemented in Apache Spark version 2.3.1. Note that finding the 
frequent itemsets in $F$ requires much less memory than in $C$ since 
$|F| << |C|$. We implemented step 2 using a single pass on $C$, which 
does not raise any memory issue since $|I|$ is small.
To obtain a limited set of rules, we then select the 
Association Rules $x \Rightarrow y$ such that $x \in I$ and:
\[
\left.
\begin{cases}
\mathrm{IR(x \Rightarrow y)} \geq \mathrm{min\_IR} \\
\mathrm{size}(x \Rightarrow y) \leq 2 \\
\mathrm{size}(x \Rightarrow y) =2 \Rightarrow \exists r \in R_1, r \prec_{\Delta\mathrm{IR}} (x \Rightarrow y)
\end{cases}
\right.
\]
$\mathrm{size}(x \Rightarrow y)$ is the size of the rule, i.e., the 
number of elements in $x$. $R_i$ is the set of rules of size i and 
$\prec_{\Delta\mathrm{IR}}$ is a partial order on $R = \cup_i{R_i}$ 
defined as follows:
\[
\forall r_1, r_2 \in R_1 \times R_2, r_1=(x_1 \Rightarrow y), r_2=(x_2, x_3 \Rightarrow y):
\]
\[
r_1 \prec_{\Delta IR} r_2 \Leftrightarrow
\begin{cases}
\mathrm{IR}(r_2) - \mathrm{IR}(r_1) \geq \Delta IR \\
x_1 = x_2 \quad \mathrm{or} \quad x_1 = x_3
\end{cases}
\]

\section{Results}
\label{sec:results}

\subsection{Classification results}

\begin{figure*}
\centering
\begin{subfigure}{\columnwidth}
  \includegraphics[width=\columnwidth]{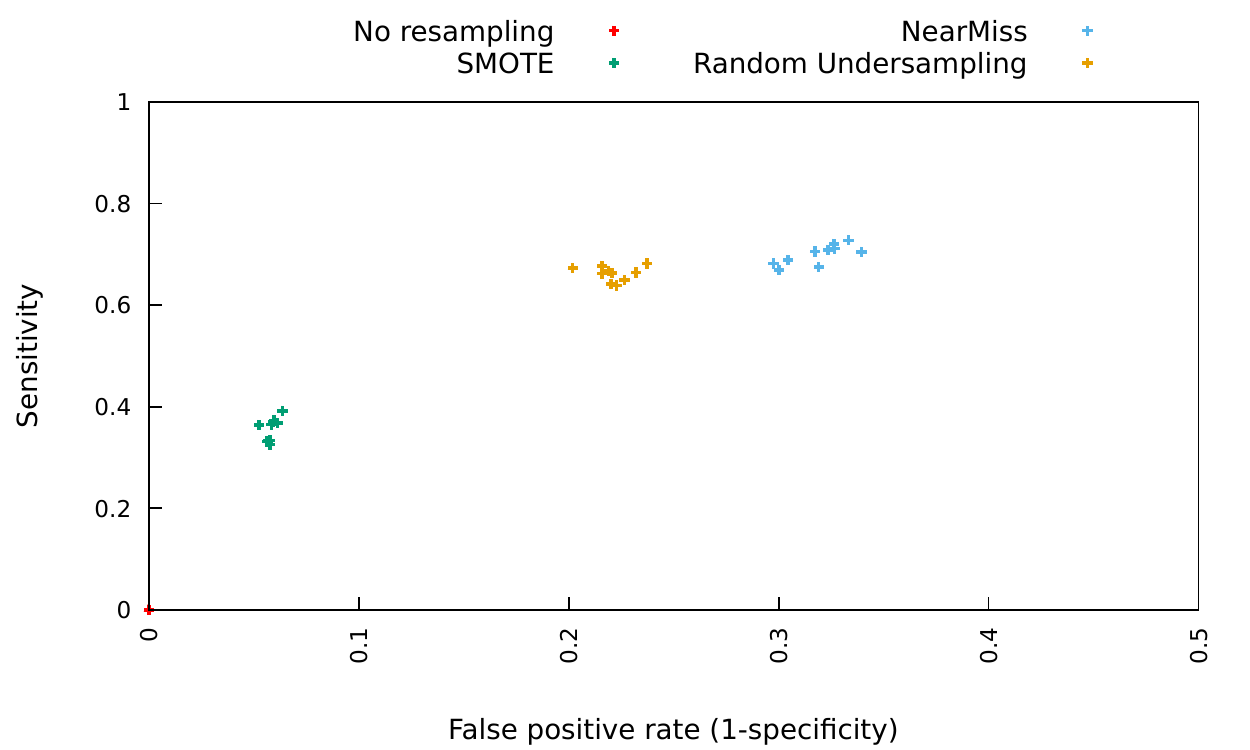}
  \caption{Customer not at home (NAH)}
  \label{fig:roc-type1}
\end{subfigure}
\begin{subfigure}{\columnwidth}
  \includegraphics[width=\columnwidth]{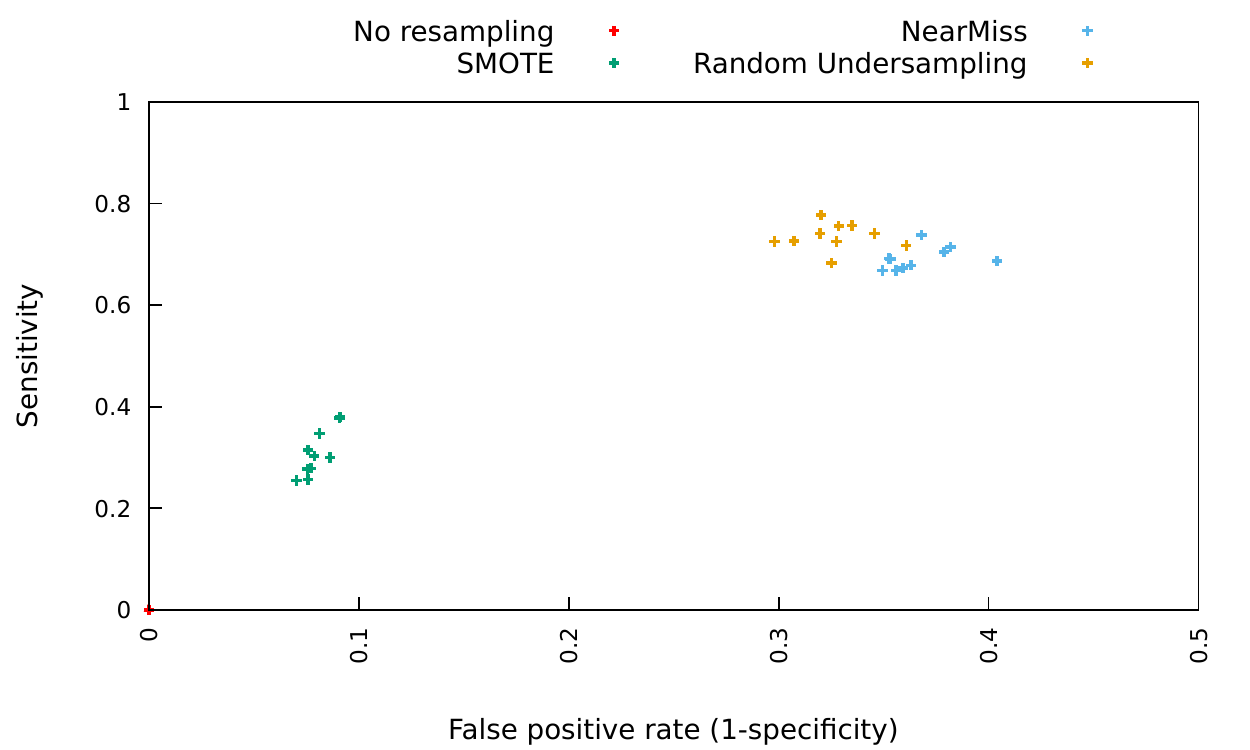}
  \caption{Stop rescheduled (SR)}
  \label{fig:roc-type4}
\end{subfigure}
\begin{subfigure}{\columnwidth}
  \includegraphics[width=\columnwidth]{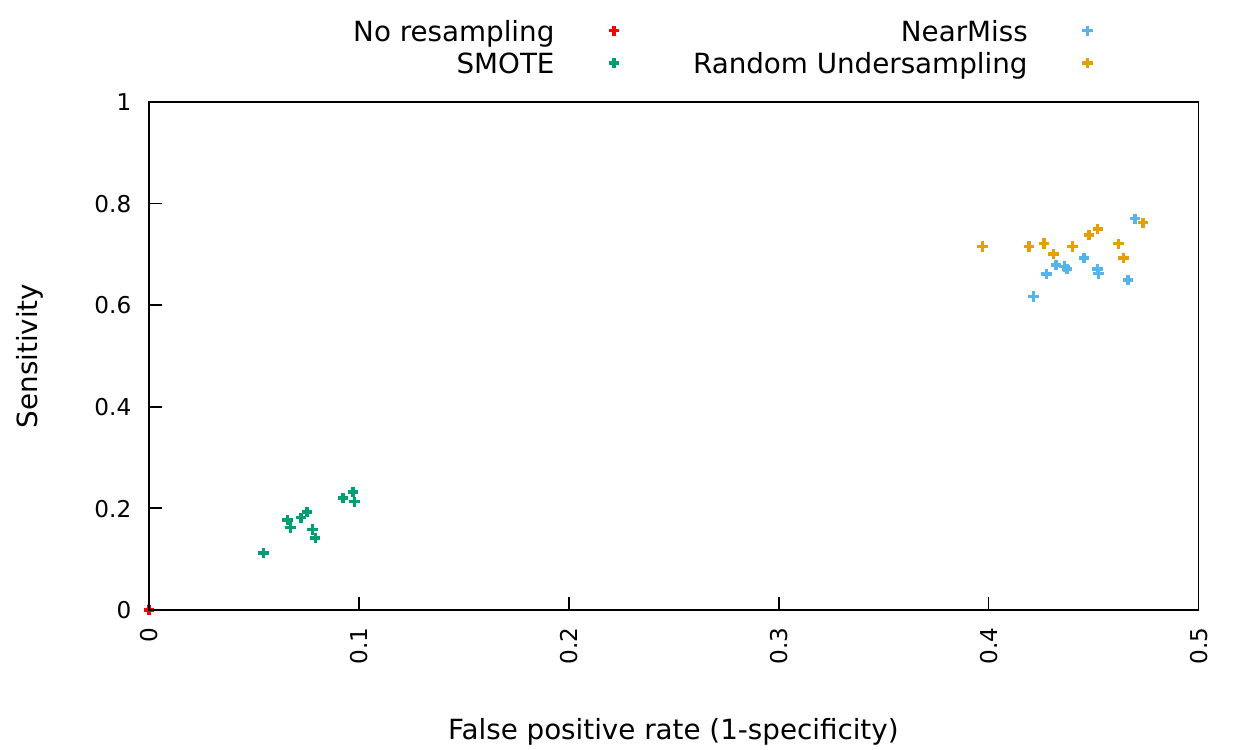}
  \caption{Refused by customer (RC)}
  \label{fig:roc-type2}
\end{subfigure}
\begin{subfigure}{\columnwidth}
  \includegraphics[width=\columnwidth]{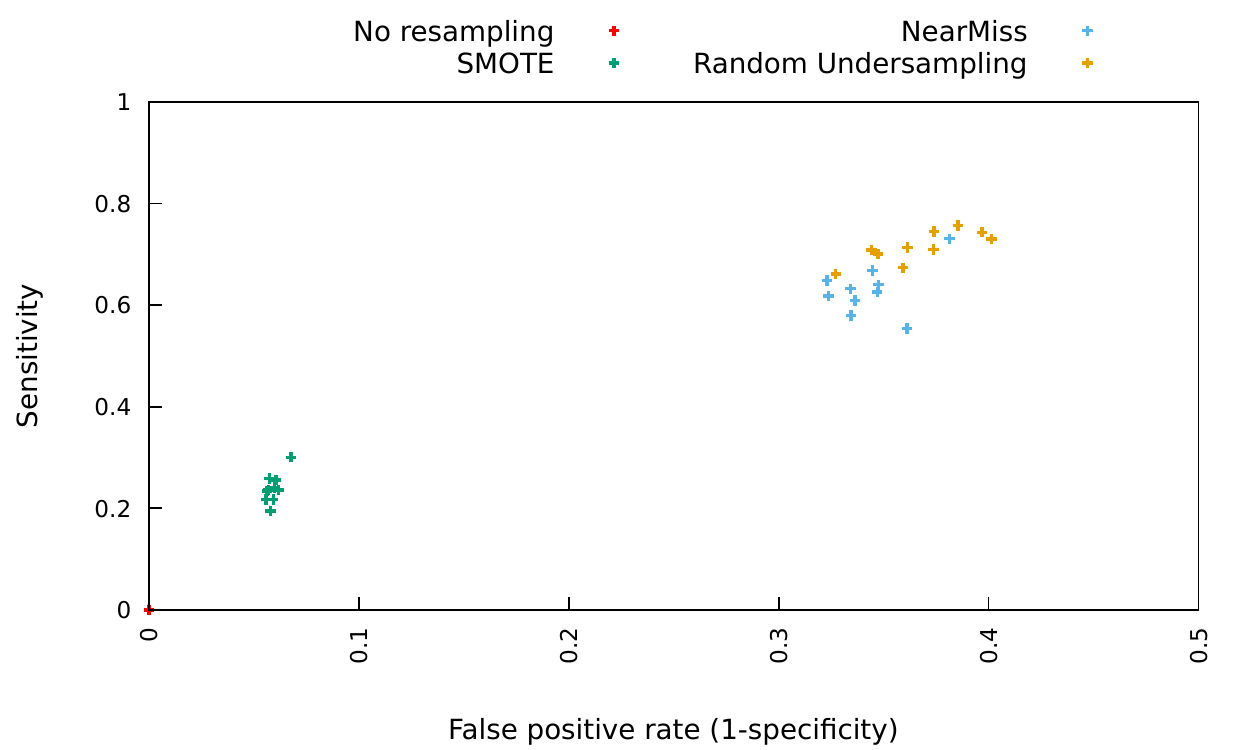}
  \caption{Canceled by customer (CC)}
  \label{fig:roc-type3}
\end{subfigure}
\begin{subfigure}{\columnwidth}
  \includegraphics[width=\columnwidth]{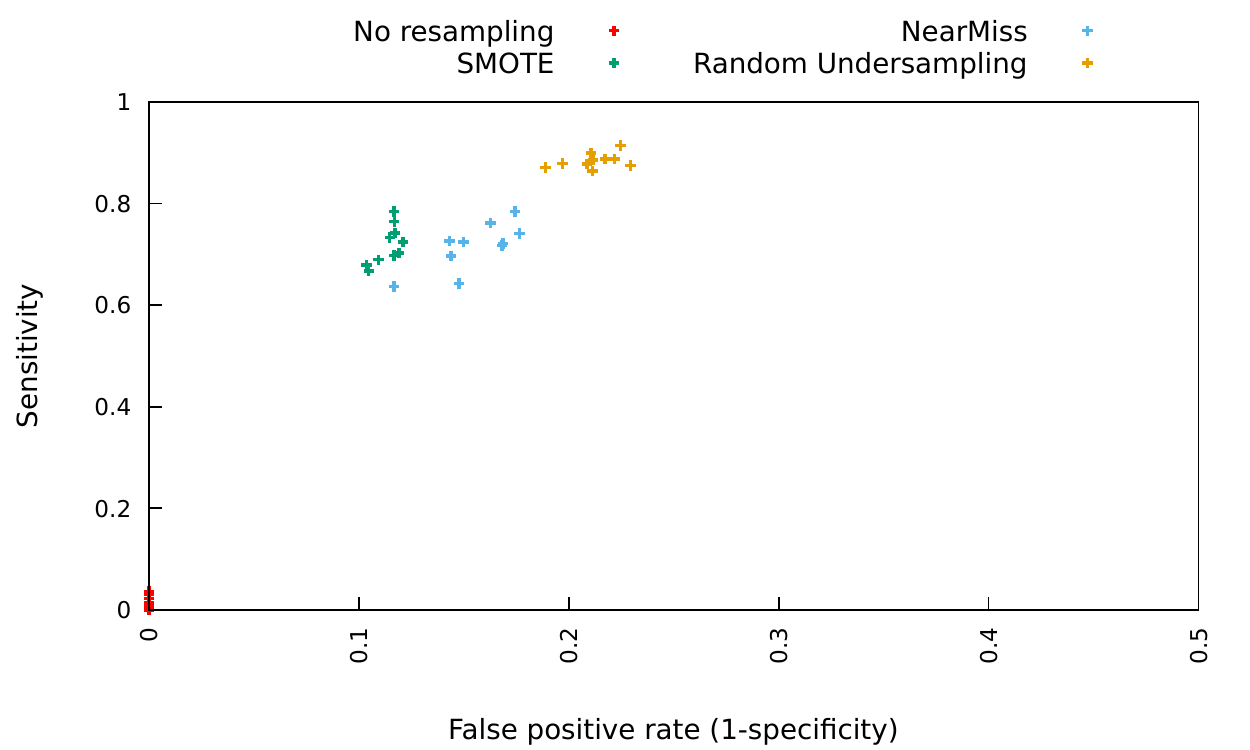}
  \caption{Not in Stock (NS)}
  \label{fig:roc-type5}
\end{subfigure}
\begin{subfigure}{\columnwidth}
\begin{tabular}{ccc}
\hline
& Sensitivity & Specificity \\
\hline
No resampling & 0 & 1\\
SMOTE & 0.36& 0.92 \\
NearMiss & 0.68 & 0.67 \\
Random Undersampling & 0.74 & 0.69\\
\hline
\end{tabular}
\caption{Average performance of the resampling methods}
\label{table:summary-stats}
\end{subfigure}
\hfill
\caption{Performance of the classifier for different types of failures and resampling methods.}
\label{fig:classification-results}
\end{figure*}

Figure~\ref{fig:classification-results} shows the sensitivity (ratio of 
true positives) and specificity (1 - ratio of false positives)
obtained for the different failure types and resampling methods. 
Without resampling, the sensitivity to failure remains 0, as expected 
in such an imbalanced dataset. Oversampling with SMOTE improves the sensitivity 
to an average of 0.36 while maintaining a high specificity of 0.92. 
Undersampling with NearMiss and Random Undersampling further increases 
the sensitivity to an average of about 0.7, with a specificity close to 
0.7. This appears to be the best compromise between sensitivity and 
specificity. On average, Random Undersampling performs slightly better 
than NearMiss. In the remainder, we focus on results obtained with 
Random Undersampling.

The classification performance is quite stable across failure types. 
With Random Undersampling, the best specificity values are obtained for 
NAH and NS, while RC is slightly under average. Sensivitiy values are 
close to average for all failure types, NS being slightly above.

\subsection{Important features and Association Rules}

\subsubsection{Customer not at home (NAH)}

\begin{figure}
\centering
\begin{subfigure}{\columnwidth}
  \includegraphics[width=\columnwidth]{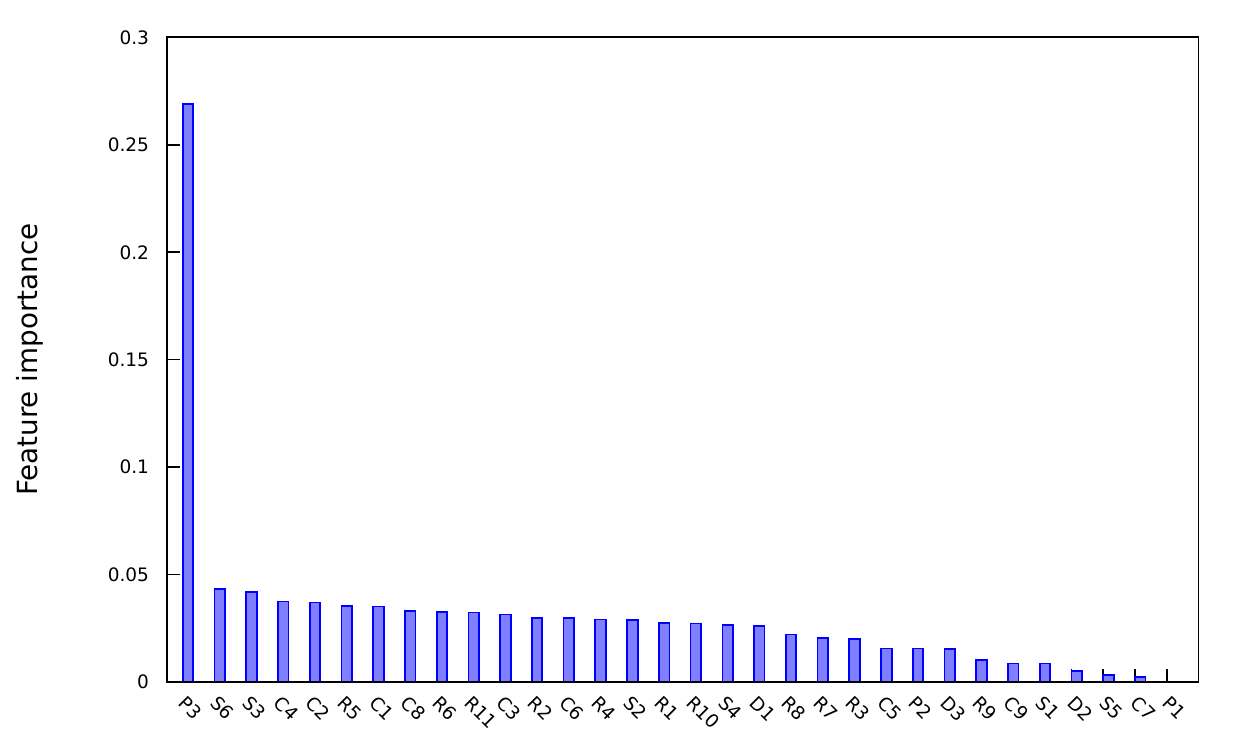}
  \caption{Feature importance}
  \label{fig:fi-nah}
\end{subfigure}
\begin{subfigure}{\columnwidth}
  \tiny
  \centering
  \begin{tabular}{cllcc}
Id & Rule & conf (\%) & IR\\
\hline
\rowcolor{rowred} 1 &  \textbf{(P3)} \IdOutboundCallAttemptResult\_V3 & 3.6 & 2.45 \\
\rowcolor{rowred} 2 & \quad \quad \textbf{(S6)} \EstimatedJobTime\_D2 & 6.0 & 4.07 \\
\rowcolor{rowred} 3 & \quad \quad \textbf{(P2)} \IdOutboundCallStatus\_V5 & 5.4 & 3.67 \\
\rowcolor{rowred} 4 & \quad \quad \textbf{(S2)} \IdCompany\_V3 & 4.6 & 3.12 \\
\rowcolor{rowred} 5 & \quad \quad \textbf{(S3)} \Volumecf\_D1 & 4.6 & 3.12 \\
\rowcolor{rowred} 6 & \quad \quad \textbf{(C9)} \IdZone\_V22 & 4.0 & 2.77 \\
\rowcolor{rowred} 7 & \quad \quad \textbf{(D3)} \Day\_V4 & 4.0 & 2.73 \\
\rowcolor{rowred} 8 & \quad \quad \textbf{(S2)} \IdCompany\_V39 & 4.0 & 2.72 \\
\rowcolor{rowred} 9 & \quad \quad \textbf{(D3)} \Day\_V2 & 4.0 & 2.72 \\
\rowcolor{rowred} 10 & \quad \quad \textbf{(R7)} \TimeWindowPickupStartTime\_D1 & 3.9 & 2.68 \\
\rowcolor{rowred} 11 & \quad \quad \textbf{(D3)} \Day\_V3 & 3.9 & 2.65 \\
\rowcolor{rowred} 12 & \quad \quad \textbf{(D2)} \Timeofday\_V2 & 3.8 & 2.61 \\
\hline
\rowcolor{rowgreen} 13 &  \textbf{(P3)} \IdOutboundCallAttemptResult\_V2 & 0.7 & 2.06 \\
\rowcolor{rowgreen} 14 & \quad \quad \textbf{(S4)} \Weightkg\_D1 & 0.6 & 2.4 \\
\rowcolor{rowgreen} 15 & \quad \quad \textbf{(C5)} \AptUnitlabel\_D1 & 0.7 & 2.22 \\
\rowcolor{rowgreen} 16 & \quad \quad \textbf{(R9)} \TimeWindowSize\_D2 & 0.7 & 2.17 \\
\end{tabular}

  \caption{Association Rules filtered with s=0.1, min\_IR=1.4, $\Delta \mathrm{IR}$=0.1.}
  \label{fig:Rules-nah}
\end{subfigure}
\caption{Customer not at home (NAH)}
\label{fig:results-nah}
\end{figure}

Figure~\ref{fig:fi-nah} shows the feature importance resulting from 
the classification of failures of type ``Customer not at home". Feature labels 
refer to the ones in Table~\ref{table:used-features}, and feature 
importance is computed as explained in Section~\ref{sec:random-forest}.
Feature importance is largely dominated by a single feature, 
\IdOutboundCallAttemptResult{ }(P3), peaking at an importance of 0.27. The next 
18 features have similar importance values, ranging from 0.026 to 0.05. The 
remaining 13 features are between 0.00 and 0.022.

Figure~\ref{fig:Rules-nah} shows the antecedents of the Association 
Rules with consequent FAIL\_NAH, selected as described in 
Section~\ref{sec:association-Rules}, with their confidence and interest 
ratio (IR). For clarity, the elements of rules of size 1 are omitted in 
rules of size 2. For instance, Rule 1 means that 
(\IdOutboundCallAttemptResult\_V3 $\Rightarrow$ FAIL\_NAH) has a 
confidence of 0.036 and an interest ratio of 2.45, and Rule 2 means that 
(\IdOutboundCallAttemptResult\_V3, \EstimatedJobTime\_D2 $\Rightarrow$ 
FAIL\_NAH) has a confidence of 0.06 and an interest ratio of 4.07. 
Rules with $\phi\geq1$ are represented in red, and rules with $\phi<1$ 
are in green. For instance, Rule 1, shown in \emph{red}, means that 
\IdOutboundCallAttemptResult=3 \emph{increases} failure probability by a 
factor of 2.45, while Rule 13, shown in  \emph{green}, means that 
\IdOutboundCallAttemptResult=2 \emph{decreases} failure probability by a 
factor of 2.06.

The rules in Figure~\ref{fig:Rules-nah} are consistent with the 
features importance in Figure~\ref{fig:fi-nah}, 
\IdOutboundCallAttemptResult{ }(P3) being the most important feature. They 
show that P3=3, which means that a call landed on voicemail, increases the 
failure probability by 2.45 times (Rule 1). This ratio increases to 3.67 if the 
call was marked failed (Rule 3: 
P2=5), to 4.07 if the estimated service time is shorter than 8 minutes (Rule 2: S6 in 
D2 (240-480]), or to 3.12 if the item volume is low (Rule 5: 
S3 in D1 [0.0, 0.002]), perhaps because less voluminous items are cheaper on average and 
customers give less value to them. The failure probability also 
increases if the item is delivered by specific companies (Rule 4: S2=3 
and Rule 8: S2=39), in the Montreal/Laval area (Rule 6: C9=22), on a Tuesday, 
Wednesday or Thursday (Rule 9: D3=2, Rule 11: D3=3, Rule 7: D3=4), if 
the time window starts between 6am and 8am (Rule 10: R7 in D1 (359.99, 480.0]), or if the 
service is planned between 10am and 12pm (Rule 12: D2=2).

Conversely, P3=2, which means that a call was answered by a human, 
reduces the failure probability by 2.06 times (Rule 13). Failure 
probability is further reduced if the item is lighter than 2~lbs (Rule 14: S4 in 
D1 [0.0, 2.0]), perhaps because drivers can leave small items unattended at the 
customer's door when they agreed during a phone call. Finally, 
failure probability is also reduced if the address has no apartment 
number (Rule 15: C5 in D1) or if the time window provided to the customer 
is short (Rule 16: R9 in D2 (120.0, 180.0]).

\subsubsection{Stop rescheduled (SR)}
\begin{figure}[t]
\centering
\begin{subfigure}{\columnwidth}
  \includegraphics[width=\columnwidth]{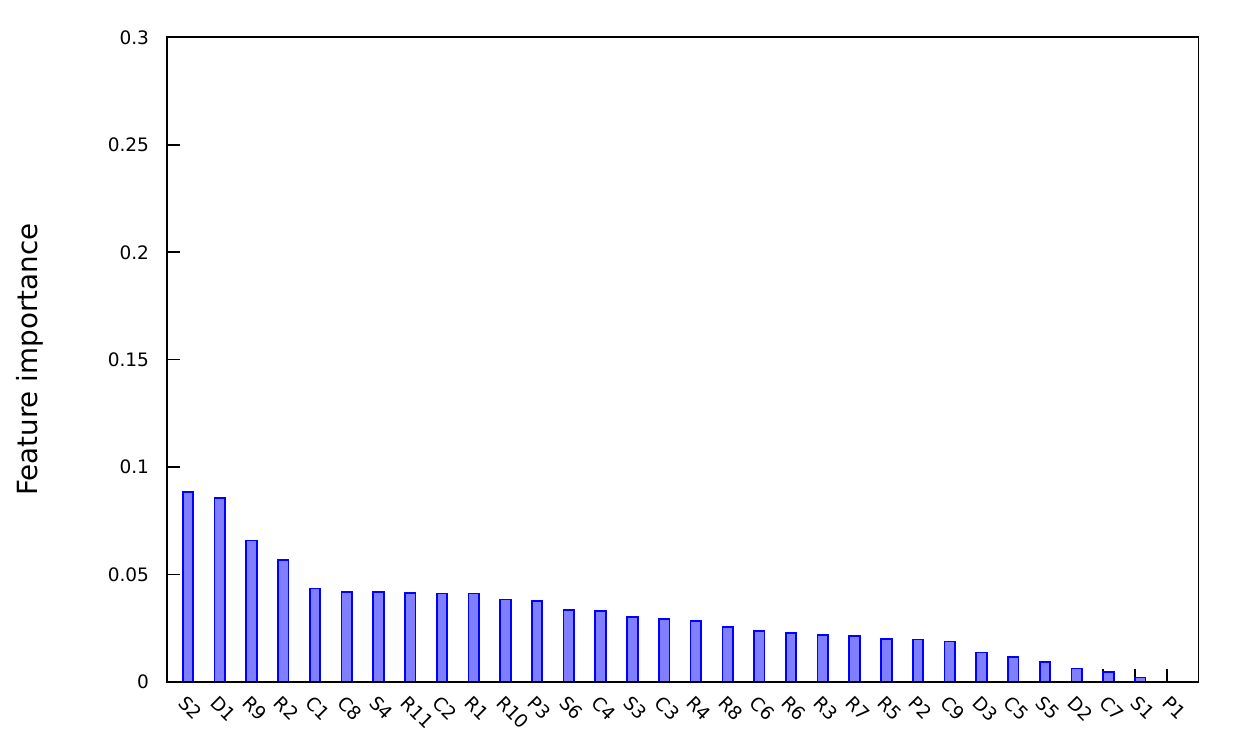}
  \caption{Feature importance}
  \label{fig:fi-sr}
\end{subfigure}
\begin{subfigure}{\columnwidth}
  \tiny
  \centering
  \begin{tabular}{cllcc}
Id & Rule & conf (\%) & IR\\
\hline
\rowcolor{rowred} 1 &  \textbf{(S2)} \IdCompany\_V39 & 1.5 & 1.88 \\
\rowcolor{rowred} 2 & \quad \quad \textbf{(P3)} \IdOutboundCallAttemptResult\_V3 & 2.5 & 3.16 \\
\rowcolor{rowred} 3 & \quad \quad \textbf{(S4)} \Weightkg\_D5 & 1.9 & 2.41 \\
\rowcolor{rowred} 4 & \quad \quad \textbf{(D3)} \Day\_V5 & 1.8 & 2.3 \\
\rowcolor{rowred} 5 & \quad \quad \textbf{(D2)} \Timeofday\_V4 & 1.7 & 2.07 \\
\rowcolor{rowred} 6 & \quad \quad \textbf{(C7)} \IdProvince\_V2 & 1.7 & 2.06 \\
\rowcolor{rowred} 7 & \quad \quad \textbf{(C9)} \IdZone\_V209 & 1.6 & 2.05 \\
\rowcolor{rowred} 8 & \quad \quad \textbf{(P2)} \IdOutboundCallStatus\_V5 & 1.6 & 2.01 \\
\hline
\rowcolor{rowred} 9 &  \textbf{(R10)} \StartSlack\_D9 & 1.4 & 1.8 \\
\rowcolor{rowred} 10 & \quad \quad \textbf{(C9)} \IdZone\_V209 & 1.6 & 1.95 \\
\rowcolor{rowred} 11 & \quad \quad \textbf{(P2)} \IdOutboundCallStatus\_V5 & 1.6 & 1.94 \\
\rowcolor{rowred} 12 & \quad \quad \textbf{(C7)} \IdProvince\_V2 & 1.6 & 1.93 \\
\hline
\rowcolor{rowgreen} 13 &  \textbf{(S2)} \IdCompany\_V3 & 0.5 & 1.78 \\
\hline
\rowcolor{rowred} 14 &  \textbf{(R11)} \EndSlack\_D6 & 1.4 & 1.77 \\
\rowcolor{rowred} 15 & \quad \quad \textbf{(C9)} \IdZone\_V209 & 1.6 & 1.97 \\
\rowcolor{rowred} 16 & \quad \quad \textbf{(C7)} \IdProvince\_V2 & 1.6 & 1.95 \\
\hline
\rowcolor{rowred} 17 &  \textbf{(R9)} \TimeWindowSize\_D3 & 1.4 & 1.77 \\
\rowcolor{rowred} 18 & \quad \quad \textbf{(P3)} \IdOutboundCallAttemptResult\_V3 & 2.4 & 3.03 \\
\rowcolor{rowred} 19 & \quad \quad \textbf{(S4)} \Weightkg\_D5 & 1.9 & 2.38 \\
\rowcolor{rowred} 20 & \quad \quad \textbf{(D3)} \Day\_V5 & 1.8 & 2.27 \\
\rowcolor{rowred} 21 & \quad \quad \textbf{(C1)} \Longitude\_D3 & 1.7 & 2.09 \\
\rowcolor{rowred} 22 & \quad \quad \textbf{(R7)} \TimeWindowPickupStartTime\_D1 & 1.7 & 2.07 \\
\rowcolor{rowred} 23 & \quad \quad \textbf{(R8)} \TimeWindowPickupEndTime\_D2 & 1.7 & 2.05 \\
\rowcolor{rowred} 24 & \quad \quad \textbf{(C2)} \Latitude\_D3 & 1.5 & 1.92 \\
\rowcolor{rowred} 25 & \quad \quad \textbf{(S3)} \Volumecf\_D3 & 1.5 & 1.91 \\
\rowcolor{rowred} 26 & \quad \quad \textbf{(P2)} \IdOutboundCallStatus\_V5 & 1.5 & 1.9 \\
\rowcolor{rowred} 27 & \quad \quad \textbf{(R10)} \StartSlack\_D9 & 1.5 & 1.89 \\
\rowcolor{rowred} 28 & \quad \quad \textbf{(C2)} \Latitude\_D2 & 1.5 & 1.88 \\
\hline
\rowcolor{rowgreen} 29 &  \textbf{(R9)} \TimeWindowSize\_D2 & 0.5 & 1.74 \\
\rowcolor{rowgreen} 30 & \quad \quad \textbf{(P3)} \IdOutboundCallAttemptResult\_V2 & 0.3 & 2.3 \\
\rowcolor{rowgreen} 31 & \quad \quad \textbf{(C7)} \IdProvince\_V1 & 0.4 & 1.98 \\
\hline
\rowcolor{rowred} 32 &  \textbf{(S4)} \Weightkg\_D5 & 1.4 & 1.74 \\
\rowcolor{rowred} 33 & \quad \quad \textbf{(P2)} \IdOutboundCallStatus\_V5 & 1.5 & 1.84 \\
\hline
\rowcolor{rowred} 34 &  \textbf{(P3)} \IdOutboundCallAttemptResult\_V3 & 1.4 & 1.7 \\
\rowcolor{rowred} 35 & \quad \quad \textbf{(P2)} \IdOutboundCallStatus\_V5 & 2.0 & 2.54 \\
\rowcolor{rowred} 36 & \quad \quad \textbf{(S6)} \EstimatedJobTime\_D2 & 1.7 & 2.06 \\
\rowcolor{rowred} 37 & \quad \quad \textbf{(C9)} \IdZone\_V209 & 1.6 & 2.01 \\
\rowcolor{rowred} 38 & \quad \quad \textbf{(C7)} \IdProvince\_V2 & 1.5 & 1.91 \\
\rowcolor{rowred} 39 & \quad \quad \textbf{(D2)} \Timeofday\_V1 & 1.5 & 1.82 \\
\hline
\rowcolor{rowred} 40 &  \textbf{(R10)} \StartSlack\_D8 & 1.3 & 1.67 \\
\rowcolor{rowred} 41 & \quad \quad \textbf{(P2)} \IdOutboundCallStatus\_V5 & 1.4 & 1.79 \\
\hline
\rowcolor{rowred} 42 &  \textbf{(S3)} \Volumecf\_D3 & 1.3 & 1.6 \\
\rowcolor{rowred} 43 & \quad \quad \textbf{(S2)} \IdCompany\_V39 & 1.5 & 1.88 \\
\hline
\rowcolor{rowred} 44 &  \textbf{(R11)} \EndSlack\_D7 & 1.2 & 1.54 \\
\rowcolor{rowred} 45 & \quad \quad \textbf{(P2)} \IdOutboundCallStatus\_V5 & 1.3 & 1.68 \\
\rowcolor{rowred} 46 & \quad \quad \textbf{(R9)} \TimeWindowSize\_D3 & 1.3 & 1.67 \\
\hline
\rowcolor{rowred} 47 &  \textbf{(S2)} \IdCompany\_V149 & 1.2 & 1.51 \\
\rowcolor{rowred} 48 & \quad \quad \textbf{(C9)} \IdZone\_V209 & 1.5 & 1.87 \\
\rowcolor{rowred} 49 & \quad \quad \textbf{(R9)} \TimeWindowSize\_D3 & 1.5 & 1.84 \\
\rowcolor{rowred} 50 & \quad \quad \textbf{(C7)} \IdProvince\_V2 & 1.4 & 1.72 \\
\rowcolor{rowred} 51 & \quad \quad \textbf{(P2)} \IdOutboundCallStatus\_V5 & 1.3 & 1.66 \\
\hline
\rowcolor{rowred} 52 &  \textbf{(S3)} \Volumecf\_D2 & 1.2 & 1.48 \\
\rowcolor{rowred} 53 & \quad \quad \textbf{(S2)} \IdCompany\_V39 & 1.4 & 1.69 \\
\rowcolor{rowred} 54 & \quad \quad \textbf{(R9)} \TimeWindowSize\_D3 & 1.3 & 1.64 \\
\hline
\rowcolor{rowred} 55 &  \textbf{(R3)} \RoadOrder\_D10 & 1.1 & 1.43 \\
\hline
\rowcolor{rowred} 56 &  \textbf{(S6)} \EstimatedJobTime\_D1 & 1.1 & 1.42 \\
\rowcolor{rowred} 57 & \quad \quad \textbf{(S2)} \IdCompany\_V39 & 1.5 & 1.9 \\
\rowcolor{rowred} 58 & \quad \quad \textbf{(R9)} \TimeWindowSize\_D3 & 1.4 & 1.77 \\
\end{tabular}

  \caption{Association Rules filtered with s=0.1, min\_IR=1.4, $\Delta \mathrm{IR}$=0.1.}
  \label{fig:Rules-sr}
\end{subfigure}
\caption{Stop rescheduled (SR)}
\label{fig:results-sr}
\end{figure}

Figure~\ref{fig:fi-sr} shows the feature importance resulting from the 
classification of failures of type ``Stop rescheduled". The failure 
importance is more uniformly distributed than for NAH. Four features 
stand out: S2 (\IdCompany), D1 (Week of Year), R9 (\TimeWindowSize) and 
R2 (\IdDriver). Feature importance remains quite constant for the next 8 
features, and it seems to decrease linearly to 0 for the remaining 
features.

The Association Rules in Figure~\ref{fig:Rules-sr} confirm the 
importance of the \IdCompany: some companies increase the failure rate 
(Rule 1: S2=39, Rule 47: S2=149), and other ones reduce it (Rule 13: 
S2=3). The failure rate is also increased by high start slack times 
(Rule 9: R10 in D9 (120.0, 129.0], Rule 40: R10 in D8 (115.0, 120.0]), 
and by high end slack times (Rule 14: R11 in D6 (116.0, 120.0], Rule 
44: R11 in D7 (120.0, 128.0]). The time window size also has an effect 
on the failure rate: D3 (180.0, 240.0] increases the failure rate (Rule 
17, 46, 49, 54 and 58), while D2 (120.0, 180.0] reduces it (Rule 29). 
As for NAH, a call landing on voicemail (P3=3) increases the failure 
probability (Rule 34). Interestingly, services executed toward the end 
of the route tend to be rescheduled more often (Rule 55: R3 in D10 
(16.0, 36.0]), and so do services with a short estimated job time (Rule 
56: S6 in D1 (0.99, 240.0]). Finally, failures are also more frequent 
for services with a median weight (Rule 32: S4 in D5(382.0, 4,7016.0]) 
or for volumes lower than 18.3~cf (Rule 42: S3 in D3 (1.45, 18.3], Rule 
52: S3 in D2 (0.002, 1.45]).

\subsubsection{Refused by customer (RC)}
\begin{figure}
\centering
\begin{subfigure}{\columnwidth}
  \includegraphics[width=\columnwidth]{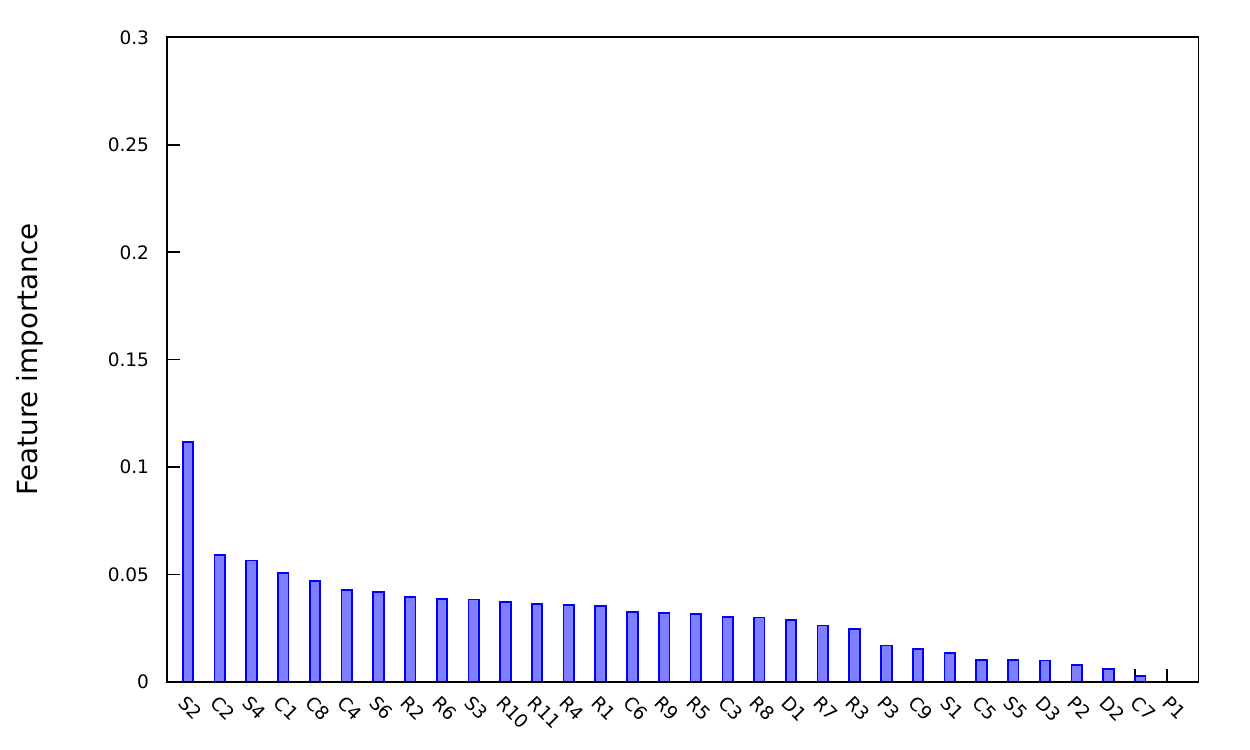}
  \caption{Feature importance}
  \label{fig:fi-rc}
\end{subfigure}
\begin{subfigure}{\columnwidth}
  \tiny
  \centering
  \begin{tabular}{cllcc}
Id & Rule & conf (\%) & IR\\
\hline
\rowcolor{rowgreen} 1 &  \textbf{(R9)} \TimeWindowSize\_D3 & 0.3 & 1.78 \\
\rowcolor{rowgreen} 2 & \quad \quad \textbf{(P3)} \IdOutboundCallAttemptResult\_V2 & 0.3 & 2.02 \\
\rowcolor{rowgreen} 3 & \quad \quad \textbf{(S5)} \IdManufacturer\_V-100.0 & 0.3 & 1.95 \\
\rowcolor{rowgreen} 4 & \quad \quad \textbf{(C5)} \AptUnitlabel\_D1 & 0.3 & 1.93 \\
\hline
\rowcolor{rowred} 5 &  \textbf{(R9)} \TimeWindowSize\_D4 & 1.0 & 1.74 \\
\rowcolor{rowred} 6 & \quad \quad \textbf{(P2)} \IdOutboundCallStatus\_V5 & 1.1 & 1.86 \\
\hline
\rowcolor{rowred} 7 &  \textbf{(S2)} \IdCompany\_V8 & 1.0 & 1.73 \\
\rowcolor{rowred} 8 & \quad \quad \textbf{(S3)} \Volumecf\_D6 & 1.3 & 2.24 \\
\rowcolor{rowred} 9 & \quad \quad \textbf{(S6)} \EstimatedJobTime\_D3 & 1.3 & 2.24 \\
\rowcolor{rowred} 10 & \quad \quad \textbf{(P2)} \IdOutboundCallStatus\_V5 & 1.1 & 1.83 \\
\hline
\rowcolor{rowred} 11 &  \textbf{(S6)} \EstimatedJobTime\_D3 & 0.9 & 1.53 \\
\rowcolor{rowred} 12 & \quad \quad \textbf{(S3)} \Volumecf\_D6 & 1.2 & 2.09 \\
\rowcolor{rowred} 13 & \quad \quad \textbf{(C7)} \IdProvince\_V2 & 1.1 & 1.84 \\
\rowcolor{rowred} 14 & \quad \quad \textbf{(S4)} \Weightkg\_D1 & 1.0 & 1.75 \\
\rowcolor{rowred} 15 & \quad \quad \textbf{(P2)} \IdOutboundCallStatus\_V5 & 1.0 & 1.66 \\
\hline
\rowcolor{rowred} 16 &  \textbf{(C1)} \Longitude\_D4 & 0.9 & 1.53 \\
\rowcolor{rowred} 17 & \quad \quad \textbf{(P2)} \IdOutboundCallStatus\_V5 & 1.0 & 1.65 \\
\hline
\rowcolor{rowred} 18 &  \textbf{(C2)} \Latitude\_D3 & 0.9 & 1.51 \\
\hline
\rowcolor{rowred} 19 &  \textbf{(R10)} \StartSlack\_D10 & 0.9 & 1.43 \\
\rowcolor{rowred} 20 & \quad \quad \textbf{(R9)} \TimeWindowSize\_D4 & 1.0 & 1.69 \\
\rowcolor{rowred} 21 & \quad \quad \textbf{(S2)} \IdCompany\_V8 & 1.0 & 1.69 \\
\rowcolor{rowred} 22 & \quad \quad \textbf{(S4)} \Weightkg\_D1 & 1.0 & 1.61 \\
\rowcolor{rowred} 23 & \quad \quad \textbf{(P2)} \IdOutboundCallStatus\_V5 & 0.9 & 1.55 \\
\hline
\rowcolor{rowred} 24 &  \textbf{(R7)} \TimeWindowPickupStartTime\_D6 & 0.9 & 1.43 \\
\rowcolor{rowred} 25 & \quad \quad \textbf{(D2)} \Timeofday\_V4 & 1.0 & 1.68 \\
\rowcolor{rowred} 26 & \quad \quad \textbf{(S4)} \Weightkg\_D1 & 0.9 & 1.58 \\
\rowcolor{rowred} 27 & \quad \quad \textbf{(P2)} \IdOutboundCallStatus\_V5 & 0.9 & 1.56 \\
\hline
\rowcolor{rowred} 28 &  \textbf{(S3)} \Volumecf\_D6 & 0.8 & 1.41 \\
\rowcolor{rowred} 29 & \quad \quad \textbf{(S4)} \Weightkg\_D1 & 1.0 & 1.61 \\
\hline
\rowcolor{rowred} 30 &  \textbf{(C9)} \IdZone\_V23 & 0.8 & 1.41 \\
\rowcolor{rowred} 31 & \quad \quad \textbf{(P2)} \IdOutboundCallStatus\_V5 & 0.9 & 1.56 \\
\hline
\rowcolor{rowred} 32 &  \textbf{(C5)} \AptUnitlabel\_D2 & 0.8 & 1.41 \\
\end{tabular}

  \caption{Association Rules filtered with s=0.1, min\_IR=1.4, $\Delta \mathrm{IR}$=0.1.}
  \label{fig:Rules-rc}
\end{subfigure}
\caption{Refused by customer (RC)}
\label{fig:results-rc}
\end{figure}

Figure~\ref{fig:fi-rc} shows the feature importance resulting from the 
classification of failures of type ``Refused by customer". Feature S2, 
\IdCompany, is standing out again. The importance seems to decrease 
linearly for the next features, with a slight increase for C2 
(\Latitude) and S4 (\Weightkg), and a slight drop between R3 and P3.

The Association Rules in Figure~\ref{fig:Rules-rc} show the effect of 
R9 (\TimeWindowSize): when in D3 (180.0, 240.0] (Rule 1), it reduces the failure rate 
by a factor of 1.78, while when in D4 (240.0, 300.0] (Rule 5), it increases it by a 
factor of 1.74. The failure rate also increases for company 8 (Rule 7: 
S2=8), for an estimated service time in D3 (480.0, 720.0] (Rule 11), in the Toronto area (Rule 16: C1 in D4 (-79.469, -79.254], Rule 
18: C2 in D3 (43.595, 43.737], Rule 30: C9=23), for the highest start 
slack times (Rule 19: R10 in D10 (129.0, 751.0]), 
for services scheduled between 11:39am and 12:08pm (Rule 24: R7 in D6 (700.0, 
728.0]), and 
for voluminous items (Rule 28: S3 in D6 (49.02, 59.6]). 

\subsubsection{Canceled by customer (CC)}
\begin{figure}
\centering
\begin{subfigure}{\columnwidth}
  \includegraphics[width=\columnwidth]{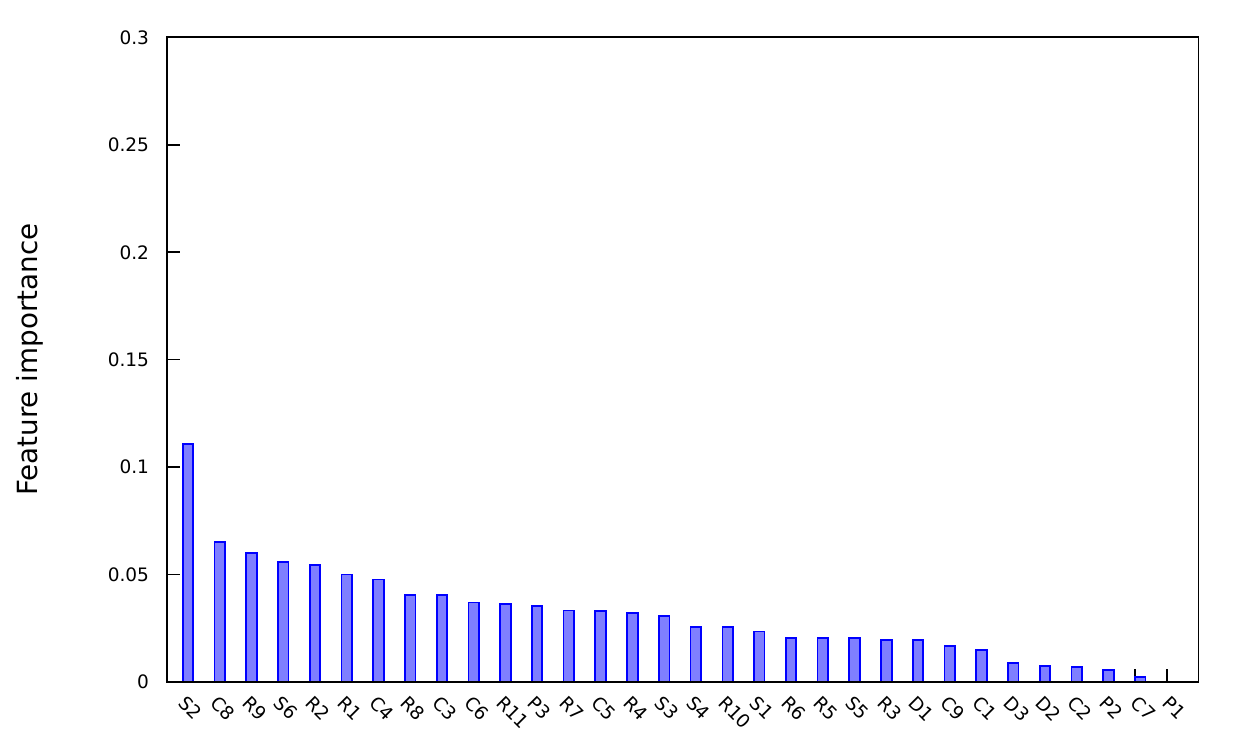}
  \caption{Feature importance}
  \label{fig:fi-cc}
\end{subfigure}
\begin{subfigure}{\columnwidth}
  \tiny
  \centering
  \begin{tabular}{cllcc}
Id & Rule & conf (\%) & IR\\
\hline
\rowcolor{rowgreen} 1 &  \textbf{(R9)} \TimeWindowSize\_D3 & 0.2 & 2.06 \\
\rowcolor{rowgreen} 2 & \quad \quad \textbf{(S5)} \IdManufacturer\_V-100.0 & 0.2 & 2.41 \\
\rowcolor{rowgreen} 3 & \quad \quad \textbf{(C5)} \AptUnitlabel\_D1 & 0.2 & 2.38 \\
\rowcolor{rowgreen} 4 & \quad \quad \textbf{(C7)} \IdProvince\_V2 & 0.2 & 2.18 \\
\hline
\rowcolor{rowred} 5 &  \textbf{(C6)} \Citylabel\_V2118 & 1.0 & 2.0 \\
\hline
\rowcolor{rowred} 6 &  \textbf{(C2)} \Latitude\_D7 & 1.0 & 2.0 \\
\rowcolor{rowred} 7 & \quad \quad \textbf{(S2)} \IdCompany\_V3 & 1.5 & 3.14 \\
\rowcolor{rowred} 8 & \quad \quad \textbf{(R9)} \TimeWindowSize\_D2 & 1.3 & 2.62 \\
\rowcolor{rowred} 9 & \quad \quad \textbf{(C9)} \IdZone\_V22 & 1.1 & 2.21 \\
\hline
\rowcolor{rowred} 10 &  \textbf{(C1)} \Longitude\_D8 & 0.9 & 1.86 \\
\rowcolor{rowred} 11 & \quad \quad \textbf{(S2)} \IdCompany\_V3 & 1.3 & 2.66 \\
\rowcolor{rowred} 12 & \quad \quad \textbf{(R9)} \TimeWindowSize\_D2 & 1.1 & 2.34 \\
\hline
\rowcolor{rowred} 13 &  \textbf{(S2)} \IdCompany\_V8 & 0.9 & 1.8 \\
\rowcolor{rowred} 14 & \quad \quad \textbf{(S3)} \Volumecf\_D6 & 1.1 & 2.25 \\
\rowcolor{rowred} 15 & \quad \quad \textbf{(C1)} \Longitude\_D4 & 1.1 & 2.24 \\
\rowcolor{rowred} 16 & \quad \quad \textbf{(S6)} \EstimatedJobTime\_D3 & 1.1 & 2.23 \\
\rowcolor{rowred} 17 & \quad \quad \textbf{(D2)} \Timeofday\_V4 & 1.0 & 2.12 \\
\rowcolor{rowred} 18 & \quad \quad \textbf{(R11)} \EndSlack\_D10 & 1.0 & 1.98 \\
\hline
\rowcolor{rowred} 19 &  \textbf{(R9)} \TimeWindowSize\_D4 & 0.9 & 1.79 \\
\rowcolor{rowred} 20 & \quad \quad \textbf{(D2)} \Timeofday\_V4 & 1.1 & 2.18 \\
\rowcolor{rowred} 21 & \quad \quad \textbf{(R11)} \EndSlack\_D10 & 0.9 & 1.91 \\
\hline
\rowcolor{rowred} 22 &  \textbf{(C1)} \Longitude\_D7 & 0.9 & 1.77 \\
\rowcolor{rowred} 23 & \quad \quad \textbf{(S2)} \IdCompany\_V3 & 1.4 & 2.81 \\
\rowcolor{rowred} 24 & \quad \quad \textbf{(R9)} \TimeWindowSize\_D2 & 1.3 & 2.56 \\
\hline
\rowcolor{rowred} 25 &  \textbf{(S2)} \IdCompany\_V3 & 0.9 & 1.74 \\
\rowcolor{rowred} 26 & \quad \quad \textbf{(S4)} \Weightkg\_D1 & 1.3 & 2.68 \\
\rowcolor{rowred} 27 & \quad \quad \textbf{(P3)} \IdOutboundCallAttemptResult\_V3 & 1.3 & 2.57 \\
\rowcolor{rowred} 28 & \quad \quad \textbf{(C9)} \IdZone\_V22 & 1.2 & 2.5 \\
\rowcolor{rowred} 29 & \quad \quad \textbf{(C7)} \IdProvince\_V1 & 1.0 & 1.97 \\
\rowcolor{rowred} 30 & \quad \quad \textbf{(R11)} \EndSlack\_D4 & 0.9 & 1.86 \\
\rowcolor{rowred} 31 & \quad \quad \textbf{(D2)} \Timeofday\_V1 & 0.9 & 1.85 \\
\hline
\rowcolor{rowred} 32 &  \textbf{(C5)} \AptUnitlabel\_D3 & 0.8 & 1.74 \\
\hline
\rowcolor{rowred} 33 &  \textbf{(C9)} \IdZone\_V22 & 0.8 & 1.6 \\
\rowcolor{rowred} 34 & \quad \quad \textbf{(P3)} \IdOutboundCallAttemptResult\_V3 & 1.4 & 2.9 \\
\rowcolor{rowred} 35 & \quad \quad \textbf{(S3)} \Volumecf\_D1 & 1.1 & 2.34 \\
\rowcolor{rowred} 36 & \quad \quad \textbf{(R9)} \TimeWindowSize\_D2 & 1.1 & 2.3 \\
\rowcolor{rowred} 37 & \quad \quad \textbf{(D3)} \Day\_V2 & 1.0 & 2.05 \\
\rowcolor{rowred} 38 & \quad \quad \textbf{(C6)} \Citylabel\_V2118 & 1.0 & 2.02 \\
\rowcolor{rowred} 39 & \quad \quad \textbf{(C1)} \Longitude\_D8 & 0.9 & 1.92 \\
\rowcolor{rowred} 40 & \quad \quad \textbf{(D3)} \Day\_V4 & 0.9 & 1.9 \\
\rowcolor{rowred} 41 & \quad \quad \textbf{(S4)} \Weightkg\_D1 & 0.9 & 1.86 \\
\rowcolor{rowred} 42 & \quad \quad \textbf{(C1)} \Longitude\_D7 & 0.9 & 1.78 \\
\rowcolor{rowred} 43 & \quad \quad \textbf{(S6)} \EstimatedJobTime\_D2 & 0.9 & 1.76 \\
\rowcolor{rowred} 44 & \quad \quad \textbf{(D2)} \Timeofday\_V4 & 0.8 & 1.74 \\
\hline
\rowcolor{rowred} 45 &  \textbf{(P3)} \IdOutboundCallAttemptResult\_V3 & 0.8 & 1.57 \\
\rowcolor{rowred} 46 & \quad \quad \textbf{(S3)} \Volumecf\_D1 & 1.2 & 2.38 \\
\rowcolor{rowred} 47 & \quad \quad \textbf{(P2)} \IdOutboundCallStatus\_V5 & 1.1 & 2.35 \\
\rowcolor{rowred} 48 & \quad \quad \textbf{(S6)} \EstimatedJobTime\_D2 & 1.1 & 2.27 \\
\rowcolor{rowred} 49 & \quad \quad \textbf{(C7)} \IdProvince\_V1 & 1.0 & 1.94 \\
\rowcolor{rowred} 50 & \quad \quad \textbf{(R9)} \TimeWindowSize\_D2 & 0.9 & 1.81 \\
\hline
\rowcolor{rowred} 51 &  \textbf{(S3)} \Volumecf\_D1 & 0.8 & 1.56 \\
\rowcolor{rowred} 52 & \quad \quad \textbf{(C7)} \IdProvince\_V1 & 0.9 & 1.94 \\
\rowcolor{rowred} 53 & \quad \quad \textbf{(R11)} \EndSlack\_D4 & 0.9 & 1.84 \\
\rowcolor{rowred} 54 & \quad \quad \textbf{(S2)} \IdCompany\_V3 & 0.8 & 1.74 \\
\rowcolor{rowred} 55 & \quad \quad \textbf{(R9)} \TimeWindowSize\_D2 & 0.8 & 1.71 \\
\rowcolor{rowred} 56 & \quad \quad \textbf{(D2)} \Timeofday\_V1 & 0.8 & 1.69 \\
\rowcolor{rowred} 57 & \quad \quad \textbf{(S4)} \Weightkg\_D1 & 0.8 & 1.69 \\
\hline
\rowcolor{rowred} 58 &  \textbf{(C5)} \AptUnitlabel\_D2 & 0.7 & 1.48 \\
\hline
\rowcolor{rowred} 59 &  \textbf{(S3)} \Volumecf\_D6 & 0.7 & 1.48 \\
\rowcolor{rowred} 60 & \quad \quad \textbf{(S6)} \EstimatedJobTime\_D3 & 1.1 & 2.18 \\
\rowcolor{rowred} 61 & \quad \quad \textbf{(C9)} \IdZone\_V209 & 1.0 & 2.08 \\
\rowcolor{rowred} 62 & \quad \quad \textbf{(C7)} \IdProvince\_V2 & 1.0 & 1.96 \\
\rowcolor{rowred} 63 & \quad \quad \textbf{(S4)} \Weightkg\_D1 & 0.8 & 1.65 \\
\hline
\rowcolor{rowred} 64 &  \textbf{(R3)} \RoadOrder\_D10 & 0.7 & 1.47 \\
\hline
\rowcolor{rowred} 65 &  \textbf{(S6)} \EstimatedJobTime\_D3 & 0.7 & 1.47 \\
\rowcolor{rowred} 66 & \quad \quad \textbf{(C7)} \IdProvince\_V2 & 0.8 & 1.71 \\
\rowcolor{rowred} 67 & \quad \quad \textbf{(S5)} \IdManufacturer\_V-100.0 & 0.8 & 1.66 \\
\rowcolor{rowred} 68 & \quad \quad \textbf{(S4)} \Weightkg\_D1 & 0.8 & 1.63 \\
\hline
\rowcolor{rowred} 69 &  \textbf{(R7)} \TimeWindowPickupStartTime\_D6 & 0.7 & 1.47 \\
\hline
\rowcolor{rowred} 70 &  \textbf{(R8)} \TimeWindowPickupEndTime\_D9 & 0.7 & 1.46 \\
\rowcolor{rowred} 71 & \quad \quad \textbf{(S4)} \Weightkg\_D1 & 0.8 & 1.6 \\
\hline
\rowcolor{rowred} 72 &  \textbf{(C2)} \Latitude\_D8 & 0.7 & 1.42 \\
\hline
\rowcolor{rowred} 73 &  \textbf{(C1)} \Longitude\_D4 & 0.7 & 1.4 \\
\end{tabular}

  \caption{Association Rules filtered with s=0.1, min\_IR=1.4, $\Delta \mathrm{IR}$=0.1.}
  \label{fig:Rules-cc}
\end{subfigure}
\caption{Canceled by customer (CC)}
\label{fig:results-cc}
\end{figure}

Figure~\ref{fig:fi-cc} shows the feature importance resulting from 
the classification of failures of type ``Canceled by customer". As in the two previous failure types,
\IdCompany{ }(S2) is standing out, and the importance seems to decrease linearly
among the other features.

The Association Rules in Figure~\ref{fig:Rules-cc} show the importance 
of \TimeWindowSize{ }(R9), as in the previous failure type: services tend 
to fail less when R9 is in D3 (180.0, 240.0] (Rule 1), and they fail 
more when R9 is in D2 (120.0, 180.0] (Rule 8, 12, 24, 36, 50 and 55) or 
in D4 (240.0, 300.0] (Rule 19). Two particular companies also have 
increased failure rates (Rule 13: S2=8, Rule 25: S2=3). The failure 
rate is also increased in the Montreal area (Rule 5: C6=2118, Rule 5: 
C2 in D7 (45.452, 45.52], Rule 72: C2 in D8 (45.52, 45.619], Rule 10: 
C1 in D8 (-73.586, -73.289], Rule 22: C1 in D7 (-73.754, -73.586], Rule 
33: C9=22), when a call lands on voicemail (Rule 45: P3=3), when the 
service is toward the end of the route (Rule 64: R3 in D10 (16.0, 
36.0]), when the service time window starts around mid-day (Rule 69: R7 
in D6 (700.0, 728.0]) or ends between 4pm and 5pm (Rule 70: R8 in 
D9 (960.0, 1020.0]), and when the item has a very low or close-to-average volume (Rule 
51: S3 in D1 [0.0, 0.002], Rule 59: S3 in D6 (49.02, 59.6]). 

\subsubsection{Not in stock (NS)} 
\begin{figure}
\centering
\begin{subfigure}{\columnwidth}
  \includegraphics[width=\columnwidth]{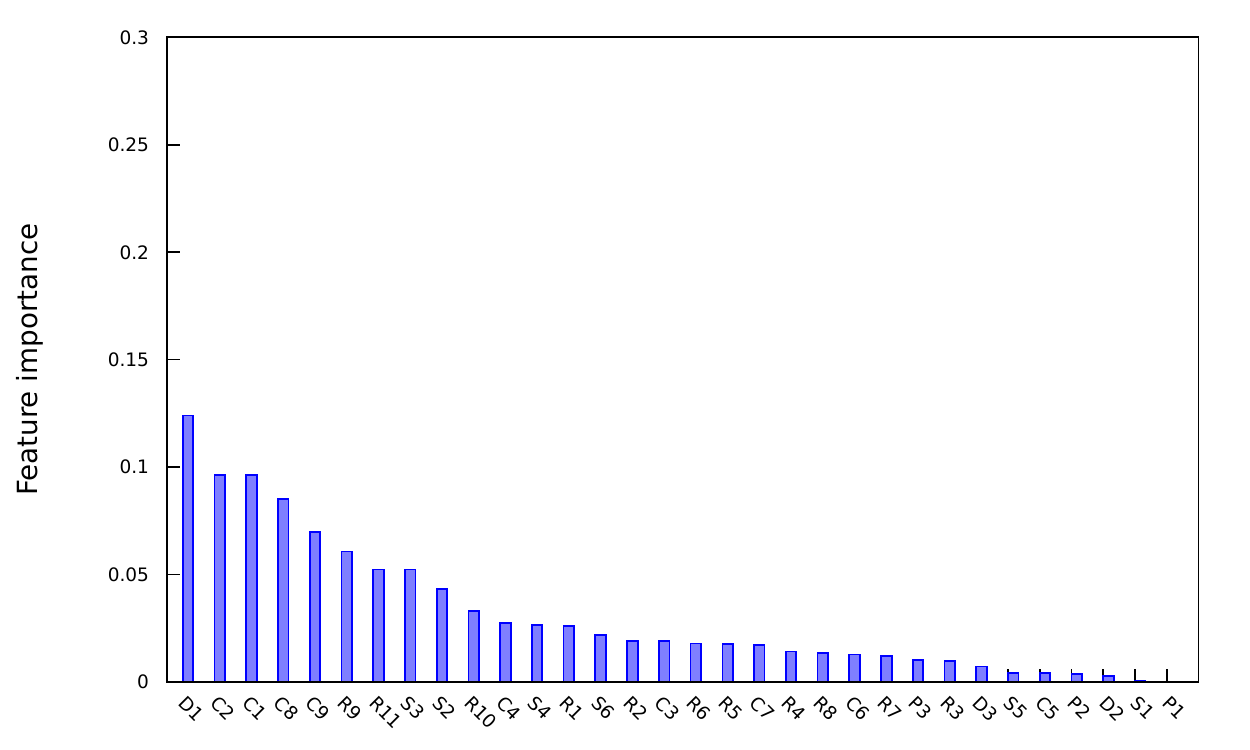}
  \caption{Feature importance}
  \label{fig:fi-ns}
\end{subfigure}
\begin{subfigure}{\columnwidth}
  \tiny
  \centering
  \begin{tabular}{cllcc}
Id & Rule & conf (\%) & IR\\
\hline
\rowcolor{rowred} 1 &  \textbf{(S2)} \IdCompany\_V158 & 10.4 & 30.81 \\
\rowcolor{rowred} 2 & \quad \quad \textbf{(C7)} \IdProvince\_V1 & 37.4 & 110.43 \\
\rowcolor{rowred} 3 & \quad \quad \textbf{(P2)} \IdOutboundCallStatus\_V5 & 11.0 & 32.4 \\
\rowcolor{rowred} 4 & \quad \quad \textbf{(S4)} \Weightkg\_D1 & 10.9 & 32.26 \\
\hline
\rowcolor{rowred} 5 &  \textbf{(R9)} \TimeWindowSize\_D1 & 1.5 & 4.4 \\
\rowcolor{rowred} 6 & \quad \quad \textbf{(S2)} \IdCompany\_V158 & 10.4 & 30.81 \\
\rowcolor{rowred} 7 & \quad \quad \textbf{(S4)} \Weightkg\_D1 & 5.9 & 17.48 \\
\rowcolor{rowred} 8 & \quad \quad \textbf{(C1)} \Longitude\_D7 & 4.9 & 14.52 \\
\rowcolor{rowred} 9 & \quad \quad \textbf{(C2)} \Latitude\_D8 & 4.2 & 12.3 \\
\rowcolor{rowred} 10 & \quad \quad \textbf{(C9)} \IdZone\_V22 & 3.5 & 10.29 \\
\rowcolor{rowred} 11 & \quad \quad \textbf{(C3)} \DoorNumberlabel\_D1 & 3.4 & 10.07 \\
\rowcolor{rowred} 12 & \quad \quad \textbf{(C1)} \Longitude\_D6 & 3.3 & 9.73 \\
\rowcolor{rowred} 13 & \quad \quad \textbf{(S6)} \EstimatedJobTime\_D3 & 3.2 & 9.32 \\
\rowcolor{rowred} 14 & \quad \quad \textbf{(C2)} \Latitude\_D9 & 3.1 & 9.19 \\
\rowcolor{rowred} 15 & \quad \quad \textbf{(C7)} \IdProvince\_V1 & 3.0 & 8.87 \\
\rowcolor{rowred} 16 & \quad \quad \textbf{(D2)} \Timeofday\_V4 & 1.9 & 5.75 \\
\rowcolor{rowred} 17 & \quad \quad \textbf{(S6)} \EstimatedJobTime\_D1 & 1.7 & 5.1 \\
\rowcolor{rowred} 18 & \quad \quad \textbf{(D3)} \Day\_V6 & 1.7 & 4.94 \\
\hline
\rowcolor{rowred} 19 &  \textbf{(D1)} \WeekofYear\_V36 & 1.2 & 3.5 \\
\rowcolor{rowred} 20 & \quad \quad \textbf{(C7)} \IdProvince\_V1 & 1.9 & 5.59 \\
\hline
\rowcolor{rowred} 21 &  \textbf{(R10)} \StartSlack\_D3 & 1.2 & 3.49 \\
\rowcolor{rowred} 22 & \quad \quad \textbf{(C9)} \IdZone\_V22 & 2.9 & 8.45 \\
\rowcolor{rowred} 23 & \quad \quad \textbf{(C3)} \DoorNumberlabel\_D1 & 2.8 & 8.22 \\
\rowcolor{rowred} 24 & \quad \quad \textbf{(S6)} \EstimatedJobTime\_D3 & 2.6 & 7.75 \\
\rowcolor{rowred} 25 & \quad \quad \textbf{(C7)} \IdProvince\_V1 & 2.4 & 7.18 \\
\rowcolor{rowred} 26 & \quad \quad \textbf{(D2)} \Timeofday\_V4 & 1.8 & 5.33 \\
\rowcolor{rowred} 27 & \quad \quad \textbf{(S6)} \EstimatedJobTime\_D1 & 1.6 & 4.76 \\
\rowcolor{rowred} 28 & \quad \quad \textbf{(S4)} \Weightkg\_D4 & 1.6 & 4.62 \\
\rowcolor{rowred} 29 & \quad \quad \textbf{(R9)} \TimeWindowSize\_D1 & 1.5 & 4.4 \\
\rowcolor{rowred} 30 & \quad \quad \textbf{(R11)} \EndSlack\_D1 & 1.5 & 4.35 \\
\rowcolor{rowred} 31 & \quad \quad \textbf{(D3)} \Day\_V6 & 1.5 & 4.29 \\
\hline
\rowcolor{rowred} 32 &  \textbf{(R11)} \EndSlack\_D1 & 1.1 & 3.35 \\
\rowcolor{rowred} 33 & \quad \quad \textbf{(C9)} \IdZone\_V22 & 2.5 & 7.38 \\
\rowcolor{rowred} 34 & \quad \quad \textbf{(C7)} \IdProvince\_V1 & 2.3 & 6.83 \\
\rowcolor{rowred} 35 & \quad \quad \textbf{(S4)} \Weightkg\_D4 & 1.5 & 4.51 \\
\rowcolor{rowred} 36 & \quad \quad \textbf{(D2)} \Timeofday\_V4 & 1.5 & 4.47 \\
\rowcolor{rowred} 37 & \quad \quad \textbf{(S6)} \EstimatedJobTime\_D1 & 1.5 & 4.4 \\
\rowcolor{rowred} 38 & \quad \quad \textbf{(S4)} \Weightkg\_D1 & 1.4 & 4.23 \\
\rowcolor{rowred} 39 & \quad \quad \textbf{(R9)} \TimeWindowSize\_D1 & 1.4 & 4.19 \\
\rowcolor{rowred} 40 & \quad \quad \textbf{(D3)} \Day\_V6 & 1.3 & 3.95 \\
\hline
\rowcolor{rowred} 41 &  \textbf{(S2)} \IdCompany\_V7 & 1.1 & 3.35 \\
\rowcolor{rowred} 42 & \quad \quad \textbf{(C1)} \Longitude\_D7 & 4.3 & 12.67 \\
\rowcolor{rowred} 43 & \quad \quad \textbf{(C9)} \IdZone\_V22 & 2.8 & 8.14 \\
\rowcolor{rowred} 44 & \quad \quad \textbf{(C7)} \IdProvince\_V1 & 2.3 & 6.81 \\
\rowcolor{rowred} 45 & \quad \quad \textbf{(S6)} \EstimatedJobTime\_D1 & 1.7 & 5.1 \\
\rowcolor{rowred} 46 & \quad \quad \textbf{(D2)} \Timeofday\_V4 & 1.6 & 4.65 \\
\rowcolor{rowred} 47 & \quad \quad \textbf{(S4)} \Weightkg\_D4 & 1.6 & 4.62 \\
\rowcolor{rowred} 48 & \quad \quad \textbf{(D3)} \Day\_V6 & 1.3 & 3.97 \\
\hline
\rowcolor{rowgreen} 49 &  \textbf{(C7)} \IdProvince\_V2 & 0.1 & 2.8 \\
\hline
\rowcolor{rowred} 50 &  \textbf{(D1)} \WeekofYear\_V44 & 0.9 & 2.7 \\
\hline
\rowcolor{rowred} 51 &  \textbf{(C1)} \Longitude\_D7 & 0.7 & 2.11 \\
\hline
\rowcolor{rowred} 52 &  \textbf{(C9)} \IdZone\_V23 & 0.7 & 2.07 \\
\rowcolor{rowred} 53 & \quad \quad \textbf{(S4)} \Weightkg\_D1 & 1.1 & 3.26 \\
\hline
\rowcolor{rowred} 54 &  \textbf{(C6)} \Citylabel\_V2118 & 0.7 & 2.06 \\
\hline
\rowcolor{rowred} 55 &  \textbf{(C1)} \Longitude\_D6 & 0.7 & 1.96 \\
\hline
\rowcolor{rowred} 56 &  \textbf{(S6)} \EstimatedJobTime\_D3 & 0.7 & 1.96 \\
\rowcolor{rowred} 57 & \quad \quad \textbf{(C3)} \DoorNumberlabel\_D1 & 2.8 & 8.35 \\
\rowcolor{rowred} 58 & \quad \quad \textbf{(C9)} \IdZone\_V22 & 1.7 & 5.14 \\
\rowcolor{rowred} 59 & \quad \quad \textbf{(C7)} \IdProvince\_V1 & 1.4 & 4.03 \\
\hline
\rowcolor{rowred} 60 &  \textbf{(C2)} \Latitude\_D7 & 0.7 & 1.95 \\
\hline
\rowcolor{rowred} 61 &  \textbf{(C9)} \IdZone\_V22 & 0.7 & 1.94 \\
\rowcolor{rowred} 62 & \quad \quad \textbf{(C3)} \DoorNumberlabel\_D1 & 4.9 & 14.54 \\
\rowcolor{rowred} 63 & \quad \quad \textbf{(S3)} \Volumecf\_D4 & 1.5 & 4.52 \\
\rowcolor{rowred} 64 & \quad \quad \textbf{(R8)} \TimeWindowPickupEndTime\_D1 & 1.1 & 3.38 \\
\rowcolor{rowred} 65 & \quad \quad \textbf{(S4)} \Weightkg\_D4 & 1.1 & 3.23 \\
\rowcolor{rowred} 66 & \quad \quad \textbf{(S6)} \EstimatedJobTime\_D1 & 1.0 & 3.06 \\
\rowcolor{rowred} 67 & \quad \quad \textbf{(D3)} \Day\_V2 & 0.9 & 2.65 \\
\rowcolor{rowred} 68 & \quad \quad \textbf{(S3)} \Volumecf\_D2 & 0.9 & 2.62 \\
\rowcolor{rowred} 69 & \quad \quad \textbf{(R11)} \EndSlack\_D2 & 0.8 & 2.51 \\
\rowcolor{rowred} 70 & \quad \quad \textbf{(P3)} \IdOutboundCallAttemptResult\_V3 & 0.8 & 2.46 \\
\hline
\rowcolor{rowgreen} 71 &  \textbf{(S2)} \IdCompany\_V3 & 0.2 & 1.93 \\
\hline
\rowcolor{rowgreen} 72 &  \textbf{(R9)} \TimeWindowSize\_D3 & 0.2 & 1.93 \\
\end{tabular}

  \caption{Association Rules filtered with s=0.1, min\_IR=1.9, $\Delta \mathrm{IR}$=0.5.}
  \label{fig:Rules-ns}
\end{subfigure}
\caption{Not in stock (NS)}
\label{fig:results-ns}
\end{figure}

Figure~\ref{fig:fi-ns} shows the 
feature importance resulting from the classification of failures of type ``Not 
in stock". The most important feature is the week of the year (D1), 
followed by features related to geographical location (C2, C1, C8 and 
C9), features related to the route (R9, R11 and R10), the company (S2) 
and the volume (S3).

This is consistent with the Association Rules in 
Figure~\ref{fig:Rules-ns}. Note that we used different filtering 
parameters for this failure type, due to the important number of 
rules with high IR. Rule 2 has an extremely high IR of 110.43, for a 
confidence of 37.4\%: it means that company 158 had a not-in-stock 
failure rate of 37.4\% in province 1 (Qu\'ebec). Some geographical 
locations spanning from Gatineau to Sorel-Tracy have increased failure rates (Rule 8, 42 and 51: C1 in D7 (-73.754, -73.586], 
Rule 12 and 55: C1 in D6 (-75.601, -73.754], Rule 9: C2 in D8 (45.52, 45.619], Rule 14: C2 in D9 (45.619, 46.328], Rule 60: 
C2 in D7 (45.452, 45.52]) while other ones have decreased failure rates (Ontario, Rule 
49: C7=2). Two specific weeks have increased failure rates: week 36 
(Rule 19), which was the week of Labor Day in 2017, and week 44 (Rule 
50), which was the week of Haloween. As for route features, time 
windows shorter than 2 hours (Rule 5: R9 in D1 [0.0, 120.0]), negative 
end slack times 
(Rule 32: R11 in D1 (-537.001, 62.0]), and start slack times between 53 
and 63 minutes (Rule 21: R10 in D3 (53.0, 63.0]) have increased failure 
rates, while time windows between 2 and 3 hours (Rule 
72: R9 in D3 (180.0, 240.0]) reduce the failure rate. Specific 
companies increase the failure 
rate (Rule 1: S2=158, Rule 41: S2=7) while other ones reduce it (Rule 
71: S2=3).

\begin{figure}
\centering
\includegraphics[width=0.92\columnwidth]{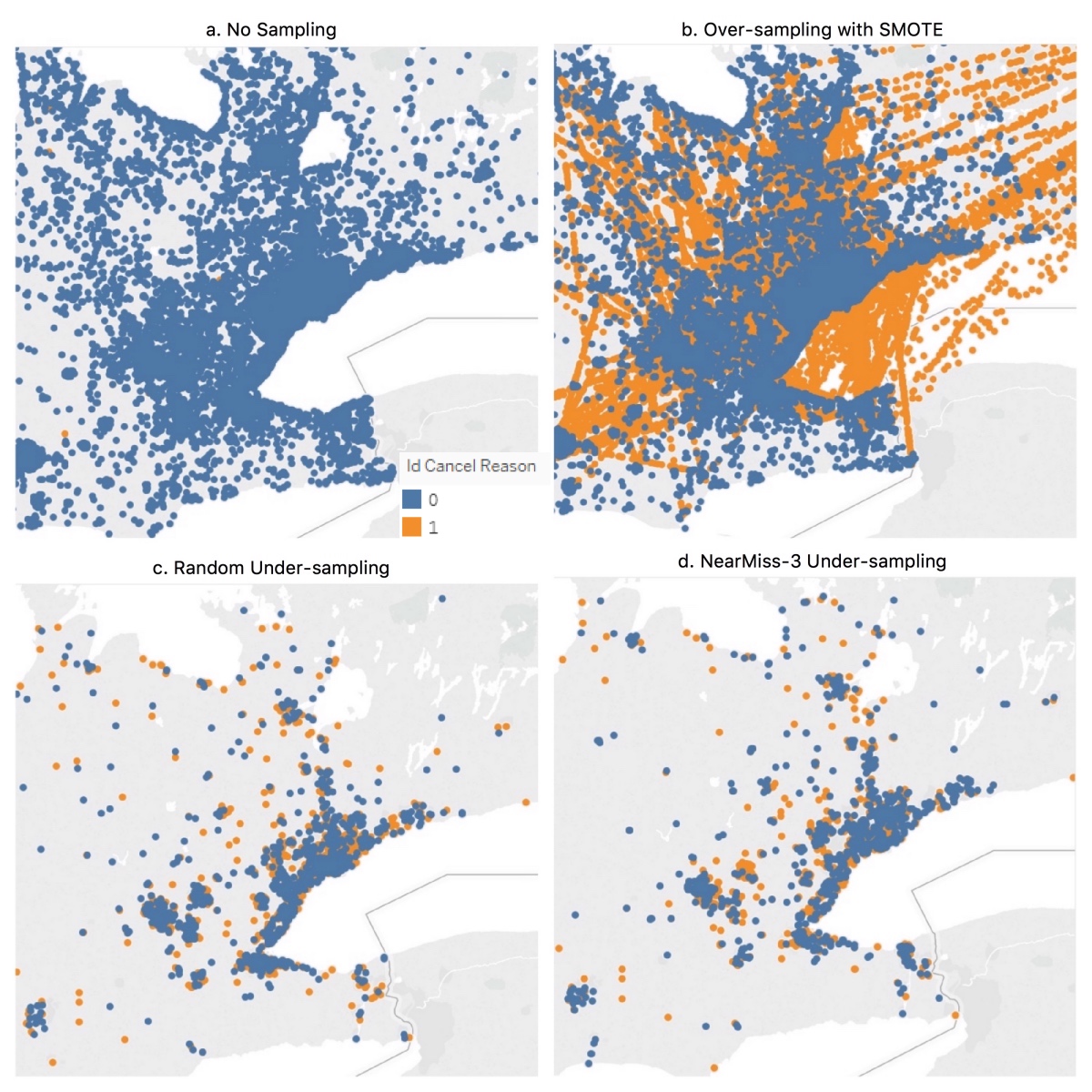}
\caption{Geographical location of some services of the training dataset after resampling with different methods.}
\label{fig:illustration-smote}
\end{figure}

\section{Discussion}

\subsection{Classification results}

Random Forests showed good performance when applied to 
the dataset pre-processed with Random Undersampling: they reach an average 
sensitivity of 0.7 and an average specificity of 0.7. Thus, 70\% of the 
failures of the studied types could be predicted, which 
represents 4,750 failed stops per year in the studied dataset. This 
prediction ability is an opportunity to save on pickup and delivery 
costs.

The classification performance could be further improved by (1) 
improving the quality of the dataset, in particular through a better 
definition and separation of failure types, (2) improving the dataset 
aggregation technique to deal with records of non-uniform sizes; our technique
essentially averages the features of the services in a stop, which 
leads to information loss, 
(3) improving the strategy to deal with dataset imbalance, perhaps 
through a more specific oversampling method.
Regarding point (3), the poor performance of SMOTE compared to the 
other resampling methods is illustrated in 
Figure~\ref{fig:illustration-smote}. 
The linear combinations of services generated by SMOTE are not realistic. In 
particular, the generated services do not respect natural boundaries 
such as lakes or uninhabitated regions, not to mention roads or actual 
addresses. This behavior is not surprising, since no such constraints 
were included in the oversampling method. Similar inconsistencies are 
also very likely to happen in other features. On the contrary, NearMiss 
and Random Undersampling maintain a realistic distribution of services, 
at the cost of reducing the dataset. A more constrained oversampling technique
might be able to address this limitation.

\subsection{Important features and Association Rules}

Overall, we observed a good agreement between the feature importance 
obtained from Random Forest and the selected Association Rules. 
Nevertheless, most Association Rules have a low confidence value, below 
5\%, which shows that failures are predicted from combinations of 
features rather than straighforward associations. We conclude that 
Association Rules, computed and selected using the methods we 
presented, are a relevant addition to RF feature importance to provide
finer-grained interpretation.

It should be noted that our extraction of Association Rules focused on 
rules where the antecedent occurs more than \emph{s} times among the 
failed stops. This explains why the selection was biased towards rules
with $\phi>1$ (rules displayed in red).

\subsection{Suggested counter-measures}

The \IdCompany{ }has a measurable effect on all the failure types. 
Specific investigations among the companies with failure rates higher than average
should be conducted, to better understand the failure causes.

Failure of type ``Customer not at home" (NAH) are very dependent on 
confirmation calls. In case such calls are not answered, additional 
ones should be scheduled, in particular if the estimated service time 
is short, if the item volume is low, if the item is not delivered on a 
Monday or Friday, if the time window starts before 8am, or if the 
service is planned between 10am and 12pm. In addition to these 
indicators, our trained Random Forest model could be used to recommend 
additional calls specifically for services predicted to fail. There 
might even be situations were multiple unanswered calls should result in the 
service to be removed from the route, if the specificity could be made 
close enough to 1.

Failures of type ``Stop rescheduled" (SR) are associated with many
features related to the route (R3, R9, R10, R11) and a few other ones related to the
type of service (S3, S4, S6). 
Such information could be included in optimizers, to facilitate the 
building of routes with less failures of this type. Start slack times 
longer than 2 hours lead to increased failure rates, which suggests 
that failures might happen due to delays in the route: dispatch centers 
might decide to skip stops when the service won't happen in the time 
window, which happens with higher probability when the start slack time 
is high. Likewise, services scheduled toward the end of the route are 
rescheduled more often than average, perhaps again due to delays in the 
route. The Time Window Size also has an effect on the failure rate: 
increased failure rates are observed for window size longer than 3 
hours. 

Failures of type ``Refused by customer" (RC) are also associated with 
route-related features (R7, R9 and R10), perhaps because delays lead 
impatient customers to refuse items. In addition, they seem to occur 
more frequently at specific geographical locations (Toronto area). 
Again, this information could be used by optimizers to build routes 
with less of such failures. Such zones might also be further 
investigated to understand the reasons for refused items. In addition, 
specific items (volume in D6 and estimated service time in D3) have an 
increased failure rate of this type, which might be reported to the 
manufacturers.

Failures of type ``Canceled by customer" (CC) are also associated with 
route-related and geographical
features (Montreal area), which could again be used by optimizers.

Finally, failures of type ``Not in Stock" (NS) are strongly related to 
one specific retailer for which more than 10\% of the 
services fail, and even 37\% in Qu\'ebec. This should 
be reported to the retailer and further investigated. 

\section*{Acknowledgement}

This work was funded by the Natural 
Sciences and Engineering Research Council of Canada (NSERC). \todo{Brigitte, is there a specific
grant number to report?}

\bibliographystyle{IEEEtran}
\bibliography{IEEEabrv,biblio}

\end{document}